\documentclass[runningheads]{llncs}

\usepackage{eccv}

\usepackage{eccvabbrv}

\usepackage{graphicx}
\usepackage{booktabs}

\usepackage[accsupp]{axessibility}  %

\usepackage{hyperref}

\usepackage{orcidlink}

\usepackage{array}
\usepackage{graphicx}
\usepackage[percent]{overpic} %

\usepackage[utf8]{inputenc} %
\usepackage[T1]{fontenc}    %

\usepackage{url}            %
\usepackage{booktabs}       %
\usepackage{amsfonts}       %
\usepackage{nicefrac}       %
\usepackage{microtype}      %
\usepackage{wrapfig}

\usepackage{tikz}
\usetikzlibrary{positioning, arrows.meta, calc}
\usepackage[dvipsnames]{xcolor}
\usepackage{adjustbox}

\usepackage{xspace}
\usepackage{siunitx}
\usepackage{graphicx}
\usepackage{caption}
\usepackage{tabu}
\usepackage{pifont}
\newcommand{\cmark}{\ding{51}\xspace}%
\usepackage{amsmath}
\usepackage{enumitem}
\usepackage{subcaption}
\usepackage{colortbl}
\usepackage{multirow}

\usepackage{listings}

\newcommand{\octicg}{\ensuremath{\mathrm{D}_8}\xspace}

\begin{document}

\title{Quick ViTs: Speeding up Vision Transformers through Equivariance} 

\titlerunning{Quick ViTs: Speeding up Vision Transformers through Equivariance}

\author{David Nordström\inst{1} \and
Johan Edstedt\inst{2} \and
Fredrik Kahl\inst{1} \and Georg Bökman\inst{3}} 

\authorrunning{Nordström et al.}

\institute{Chalmers University of Technology \and Linköping University \and University of Amsterdam\\
\url{https://github.com/davnords/octic-vits}\\
}

\maketitle

\begin{abstract}

  Natural images exhibit strong geometric regularities: local structures, such as edges, corners, and textures, appear in many orientations and mirror configurations. Since Vision Transformers (ViTs) operate on square image patches, these transformations naturally correspond to the dihedral symmetry group $\mathrm{D}_8$, also known as the octic group.
  Recent work has shown that ViTs can be made reflection equivariant and more efficient than standard ViTs simultaneously by implementing the linear layers in the Fourier domain of the reflection group. In this work, we extend the equivariance to reflections and rotations and analyze the scalability of the resulting networks. Our Quick ViTs, based on octic equivariant linear layers, achieve 5.33x reductions in FLOPs and up to 8x reductions in memory compared to ordinary linear layers. By analyzing the arithmetic intensity of these layers, we identify theoretical limits on how much the FLOP savings translate into throughput improvements on modern GPUs. However, these limitations disappear as the embedding dimensions increase. Enabled by their computational efficiency, we conduct a broader empirical evaluation of equivariant ViTs than in previous work. Upon training supervised (DeiT-III) and self-supervised (DINOv2) on ImageNet-1K, we find that our Quick ViTs match or exceed baseline accuracy while at the same time providing substantial efficiency gains.
  
\end{abstract}

\section{Introduction} \label{sec:intro}
In the pursuit of flexible yet scalable models,
Vision Transformers (ViTs)~\cite{dosovitskiy2021an} have emerged as the dominant architecture in modern computer vision. While ViTs are sometimes heralded as having minimal inductive biases, one key to their success is the construction of visual tokens from image patches and the weight-sharing over the tokens implemented by transformer layers.
This weight-sharing ensures permutation equivariance and yields better performing models than allowing arbitrary interactions over tokens~\cite{bachmann2023scaling}.

\begin{figure}[t]
\vskip 0.2in
\begin{center}
\centerline{\includegraphics[width=0.99\columnwidth]{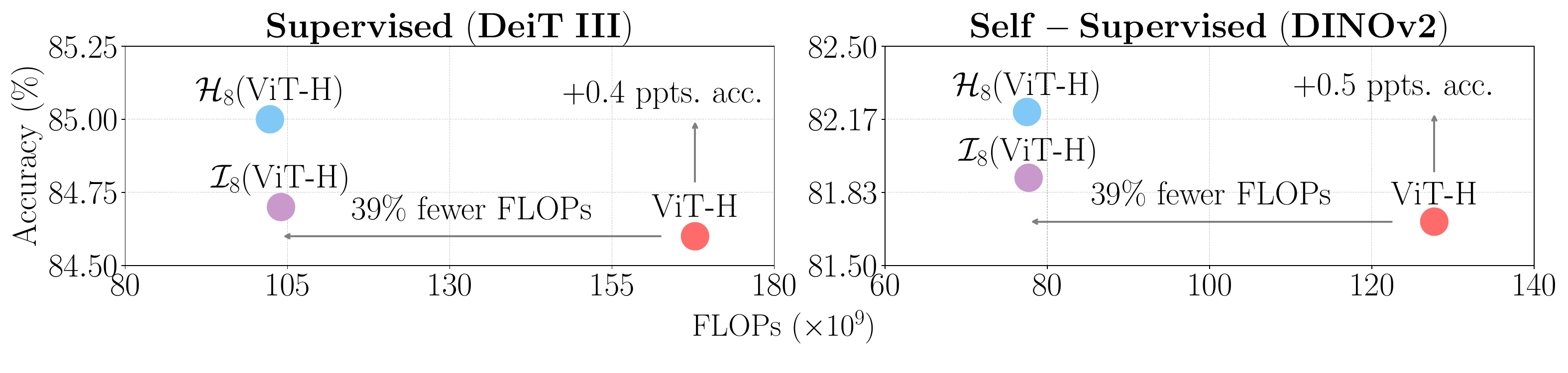}}
\caption{\textbf{Computational savings}. Using our equivariant layers in ViTs significantly reduces the computational complexity without sacrificing accuracy on ImageNet-1K, for both supervised and self-supervised training. Detailed results can be found in Section \ref{sec:experiments}.  }
\label{fig:computational_savings}
\end{center}
\vskip -0.2in
\end{figure}

In general, equivariance can provide a powerful inductive bias in neural networks by enforcing structured responses to transformations such as permutations, translations, rotations, or reflections.
However, implementations of equivariant networks often lack in computational efficiency compared to non-equivariant counterparts~\cite{klee2023a}, limiting their popularity in large scale settings.
The permutation equivariance in transformers is a clear exception to this rule, as it instead improves the computational efficiency of the model.
Recent work has furthermore demonstrated how equivariance under reflections can be implemented efficiently in ViTs, obtaining equivariant ViTs that are more computationally efficient than standard ViTs~\cite{flopping-for-flops}.
In this work, we extend the equivariance to rotations and reflections and introduce Quick ViTs, a family of computationally efficient equivariant vision transformers.
Our central observation is that implementing equivariance in the Fourier domain of the symmetry group allows the linear layers to be computed significantly more efficiently.
While \cite{flopping-for-flops} only present proof-of-concept experiments,
we provide a more extensive evaluation using a self-supervised DINOv2 pipeline and evaluating on both segmentation and classification.
We find that rotation and reflection equivariant layers can effectively preserve or improve performance of baseline models while requiring significantly fewer FLOPs, see~\Cref{fig:computational_savings}.

The analysis of computational complexity in \cite{flopping-for-flops} is limited to counting FLOPs.
We also consider the arithmetic intensity of the equivariant layers, which is crucial to obtain optimal GPU throughput.
The analysis of arithmetic intensity %
shows how the combination of hardware and datatype %
determines how much of the reduction in FLOPs can be expected to translate into throughput improvements.

In summary, our contributions are as follows:
\begin{enumerate}[topsep=0.5ex,itemsep=0ex,label=(\alph*)]
\item We introduce Quick ViTs, a family of vision transformers with octic equivariant layers (\Cref{sec:method}).
We propose two variants that are either fully equivariant ($\mathcal{I}_8$) or break equivariance in late layers of the network ($\mathcal{H}_8$).
The equivariant linear layers use $5.33$ times fewer FLOPs and $8$ times less memory per feature dimension than ordinary linear layers.

\item We analyze the arithmetic intensity of the equivariant linear layers and relate this analysis to the observed throughput improvements (\Cref{sec:comp-eff}).
\item We demonstrate empirically that our ViTs can be integrated with state-of-the-art training recipes (DeiT III and DINOv2) without re-tuning hyperparameters (\Cref{sec:experiments}). 
We achieve a $40\%$ FLOP saving with our $\mathcal{I}_8(\text{ViT-H})$ and $\mathcal{H}_8(\text{ViT-H})$ models while matching baseline performance~(\Cref{fig:computational_savings}).
\item We study the effects of different invariantization methods including symmetry breaking and varying number of equivariant layers (\Cref{sec:equi_break}).
These ablations help guide future research on equivariant architectures at scale.
\end{enumerate}

Our approach seamlessly integrates into existing ViT architectures and leads to significant gains in throughput and memory efficiency, without sacrificing accuracy. %
In addition to introducing new state-of-the-art ViTs, from a broader perspective, our contributions give clear evidence that equivariance can matter at scale, which has been a subject of debate in recent literature \cite{abramson2024accurate, wang2024swallowing,brehmer:2024,flopping-for-flops}.

\section{Related Work}  %

\subsubsection{Vision Transformers.} The ViT was introduced by~\cite{dosovitskiy2021an} and has subsequently achieved state-of-the-art results in many domains of computer vision~\cite{detr:2020, clip:2021, sam:2023, vggt:2025, edstedt2026romav2, nordstrom2026loma}. Significant efforts have been made to scale ViTs \cite{zhai2022scalingvisiontransformers, scalingvits22b:2023}, alongside strategies to do so efficiently~\cite{vitinshape:2024}.
Our approach to making ViTs faster via equivariance is orthogonal to other approaches such as pruning or quantizing.
We view it as interesting future work to combine our method with such other efficiency improvements.

\subsubsection{Equivariant Networks.} 
The equivariance of CNNs to (cyclic) image translations can be extended to incorporate larger symmetry groups such as rotations and reflections, as shown by~\cite{cohen2016group, dieleman2016exploiting} using Group Equivariant CNNs (G-CNNs). 
\cite{cohen-steerable, weiler_cesa_2019} generalized G-CNNs to steerable CNNs, where the features transform according to general group representations.
Our ViTs can be seen as a more scalable ViT analogue of the octic steerable CNNs by \cite{cohen-steerable}. There have also been prior efforts on attention- and transformer-based equivariant architectures, for both point clouds~\cite{fuchs2020se3transformers,lietransformer:2021, assaad2023vntransformer, liao2023equiformer} and, more closely to ours, images~\cite{romero2020attentive, xu20232, making-vits-shift-equi:2024, steerabletransformers:2024}.
Even when equivariance is not explicitly built into the architecture, neural networks may still learn approximately equivariant representations when trained on data with the corresponding symmetries~\cite{bokman-kahl-neurips-2023}.
However, these prior works do not obtain computational benefits over non-equivariant networks, in contrast to our ViTs. 

Our work is part of an ongoing research direction of studying and improving the scalability of equivariant networks~\cite{bekkers2024fastexpressivesenequivariant,brehmer:2024,bharadwaj2025efficient,vadgama2025utility,cuEquivariance}.
Prior work in this direction mostly focuses on point cloud data, with the notable exception of~\cite{efficient-equivariant-network:2021} and~\cite{flopping-for-flops} who consider images. \cite{efficient-equivariant-network:2021} increase the computational efficiency of G-CNNs but in contrast to our work do not achieve benefits over standard networks.

Directly inspiring our work, \cite{flopping-for-flops} demonstrate that incorporating reflection equivariance into modern image classifiers increases computational efficiency while maintaining classification performance.
In this work, we use both rotation and reflection equivariance, to obtain larger computational savings.
We evaluate our equivariant ViTs on a larger set of tasks than prior literature on equivariant image models and we extend the analysis of computational efficiency from counting FLOPs to looking at arithmetic intensity. We also analyze the optimal number of equivariant layers and invariantization.

\section{Method}
\label{sec:method}

In this section, we explain the construction of rotation and reflection equivariant ViT layers.
We begin with preliminaries %
in Section~\ref{subsec:octic-equi}, followed by the introduction of equivariant ViTs in~\Cref{subsec:octic-vits}, specifics of the transformer layers in Section~\ref{subsec:octic-layers}.
We summarize the most important notation at the end of~\Cref{subsec:notation}.

\subsection{Preliminaries on Octic Equivariance}\label{subsec:octic-equi}
In this work, we focus on the dihedral group with eight elements, \begin{equation}
\octicg = \{e,r,r^2,r^3,s,sr,sr^2,sr^3\},
\end{equation}
such that $r^4=s^2=e$ is the identity element and $r^3=srs$.\footnote{The reader is cautioned that \octicg is sometimes alternatively denoted $\mathrm{D}_4$.} 
\octicg acts on images by horizontal reflections $s$ and $\ang{90}$ anti-clockwise rotations $r$ (the reflection axis and rotation direction are arbitrary choices).
\octicg is also called the octic group, and we opt for this shorter name throughout.
We discuss the generalization to larger dihedral groups in Appendix~\ref{appendix:larger-dihedral-groups}.

In equivariant network layers, we need to keep track how the output transforms when the input transforms under \octicg.
As mentioned in the introduction, transformer layers are generally equivariant under permutations of the tokens.
When we reflect and rotate an image, the image patches get permuted but also reflected and rotated.
In our equivariant ViTs, we will specify a transformation of the channel dimension of the visual tokens that should correspond to the internal reflections/rotations of the image patches.
Then we will construct our layers such that they preserve equivariance with respect to both permutations (obtained by default in token-wise transformer layers) as well as the additional internal transformation, and the internal transformation will be formalized using a real group representation of \octicg as explained next.

In our setting, a real group representation is a vector space $\mathbb{R}^n$ equipped with a group homomorphism $\rho$ from \octicg to the group of $n\times n$ invertible real matrices. 
In other words for every $g\in \octicg$, $\rho(g)$ is an invertible matrix and for every $g,h\in \octicg$, $\rho(gh)=\rho(g)\rho(h)$. %
$\rho$ specifies how a vector in $\mathbb{R}^n$
transforms under \octicg.
There are only a few different representations of \octicg used in this work, we list them below as examples.
Finally, it is worth mentioning that all representations considered here are orthogonal, i.e., satisfy $\rho(g)^{-1}=\rho(g)^\mathsf{T}$.

The atomic building blocks of group representations are the so-called irreducible representations.
\begin{example}[Irreducible representations]
\label{ex:irreps}
   The five irreducible representations, short irreps, of \octicg are defined by
   \begin{equation}
    \begin{split}
       &\rho_{\mathrm{A1}}(r)=\rho_{\mathrm{A1}}(s)=1; \qquad\quad
       \rho_{\mathrm{A2}}(r)=1,\ \rho_{\mathrm{A2}}(s)=-1; \\
       &\rho_{\mathrm{B1}}(r)=-1,\ \rho_{\mathrm{B1}}(s)=1; \quad
       \rho_{\mathrm{B2}}(r)=\rho_{\mathrm{B2}}(s)=-1; \\
       &\text{and}\quad \rho_{\mathrm{E}}(r)=\begin{pmatrix} 0 & -1 \\ 1 & 0 \end{pmatrix},\ \rho_{\mathrm{E}}(s)=\begin{pmatrix}-1 & 0 \\ 0 & 1\end{pmatrix}.
    \end{split}
   \end{equation}
   We use the same notation as the original work on steerable CNNs \cite{cohen-steerable} for these irreps, but choose a different basis for $\rho_E$.
    It is known from elementary representation theory that any representation of \octicg can be decomposed into irreps as
\begin{equation}\label{eq:irrep_sum}
    \rho(g) = Q\left( 
    \raisebox{0.2ex}{$\bigoplus_{i\in\{\mathrm{A1}, \mathrm{A2},\mathrm{B1}, \mathrm{B2}, \mathrm{E}\}}m_i \rho_i(g)$}
    \right) Q^{-1}
\end{equation}
where $\oplus$ denotes direct sum of representations, or stacking matrices in a block diagonal, and we write $m_i \rho_i(g)$ for $\oplus$'ing $\rho_i(g)$ with itself $m_i$ times. Here
$Q$ is an invertible matrix and the $m_i$ are integers specifying the multiplicity of each irrep.
\end{example}

\begin{example}[Regular representation]
    The regular representation $\rho_\text{reg}$ 
    can be thought of as \octicg acting canonically on the vector space of functions $\phi:\octicg \to \mathbb{R}$: 
    \begin{equation}
        \left[\rho_\text{reg}(g)\phi\right](h) = \phi(g^{-1}h).
    \end{equation}
    We identify each $\phi:\octicg \to \mathbb{R}$ with the vector 
    \begin{equation}
        \begin{pmatrix}\phi(e) & \phi(r^3) & \phi(r^2) & \phi(r) & \phi(s) & \phi(sr^3) & \phi(sr^2) & \phi(sr)\end{pmatrix}^\mathsf{T}\in\mathbb{R}^8
    \end{equation}
    so that $\rho_\text{reg}(g)$ can be written as a permutation matrix.
    Importantly, $\rho_\text{reg}$ commutes with pointwise activation functions such as GELU~\cite{hendrycks2016gelu}.
\end{example}

\begin{example}[Isotypical decomposition / Fourier transform] \label{ex:isotypic}
    The regular representation $\rho_\text{reg}$ can be block-diagonalized to its isotypical decomposition $\rho_\text{iso}$ through \eqref{eq:irrep_sum} as $\rho_\text{reg}(g) = Q_\text{reg} \rho_\text{iso}(g) Q_\text{reg}^{-1}$
    with
    \begin{equation}
        \rho_\text{iso}(g) = 
            \rho_\text{A1}(g)
            \oplus \rho_\text{A2}(g)
            \oplus \rho_\text{B1}(g)
            \oplus \rho_\text{B2}(g)
            \oplus 2\rho_\text{E}(g).
    \end{equation}
    The change of basis $Q_\text{reg}$ (written out in full in~\Cref{appendix:fourier-transform})
    is the inverse Fourier transform of \octicg, with $Q_\text{reg}^{-1}=Q_\text{reg}^\mathsf{T}$ being the Fourier transform.
\end{example}

\begin{figure}[t]
    \centering
\adjustbox{max width=0.95\linewidth}{
        \trimbox{1.15cm 0 0.43cm 0}{
            \definecolor{figtextcolor}{HTML}{555555}
\definecolor{basiccolor}{HTML}{70d6ff}
\definecolor{colorA1}{HTML}{C999CC}
\definecolor{colorA2}{HTML}{E51EA3}
\definecolor{colorB1}{HTML}{ffd670}
\definecolor{colorB2}{HTML}{e9ff70}
\definecolor{colorE}{HTML}{d0f4de}
\begin{tikzpicture}[
    block/.style={draw, thick},
    label/.style={font=\Large},
    /tikz/mycolors/standard/.initial = basiccolor!40,
    /tikz/mycolors/a1/.initial = colorA1!70,
    /tikz/mycolors/a2/.initial = colorA2!20,
    /tikz/mycolors/b1/.initial = colorB1!70,
    /tikz/mycolors/b2/.initial = colorB2!70,
    /tikz/mycolors/e/.initial = colorE,
    standardstyle/.style={block, fill=\pgfkeysvalueof{/tikz/mycolors/standard}},
    a1style/.style={block, fill=\pgfkeysvalueof{/tikz/mycolors/a1}},
    a2style/.style={block, fill=\pgfkeysvalueof{/tikz/mycolors/a2}},
    b1style/.style={block, fill=\pgfkeysvalueof{/tikz/mycolors/b1}},
    b2style/.style={block, fill=\pgfkeysvalueof{/tikz/mycolors/b2}},
    estyle/.style={block, fill=\pgfkeysvalueof{/tikz/mycolors/e}},
    standardstyleweight/.style={block, fill=\pgfkeysvalueof{/tikz/mycolors/standard}},
    a1styleweight/.style={block, fill=\pgfkeysvalueof{/tikz/mycolors/a1}},
    a2styleweight/.style={block, fill=\pgfkeysvalueof{/tikz/mycolors/a2}},
    b1styleweight/.style={block, fill=\pgfkeysvalueof{/tikz/mycolors/b1}},
    b2styleweight/.style={block, fill=\pgfkeysvalueof{/tikz/mycolors/b2}},
    estyleweight/.style={block, fill=\pgfkeysvalueof{/tikz/mycolors/e}},
]
\begin{scope}
    \filldraw[standardstyleweight] (0,0) rectangle (4,4)
    node at (0, 2) [left, figtextcolor] {\small$C$};
    
    \filldraw[standardstyle] (4.3,0) rectangle (5.0,4) node[above left, figtextcolor, xshift=-1mm] {\small$L$};
\end{scope}

\draw[-{Stealth[length=3mm]}, thick, figtextcolor] (5.7, 2) -- (7.3, 2) node[midway, above, yshift=1mm] {\small Fourier domain} node[midway, below, yshift=-1mm] {\small $81\%$ sparsity};

\begin{scope}[xshift=8cm]
    \draw[thick] (0,0) rectangle (4,4);
    
    \def\sA{0.5} %
    \def\sB{0.5} %
    \def\sC{0.5} %
    \def\sD{0.5} %
    \def\sE{1.0} %
    
    \filldraw[a1styleweight] (0, 4-\sA) rectangle (\sA, 4);
    \filldraw[a2styleweight] (\sA, 4-\sA-\sB) rectangle (\sA+\sB, 4-\sA);
    \filldraw[b1styleweight] (\sA+\sB, 4-\sA-\sB-\sC) rectangle (\sA+\sB+\sC, 4-\sA-\sB);
    \filldraw[b2styleweight] (\sA+\sB+\sC, 4-\sA-\sB-\sC-\sD) rectangle (\sA+\sB+\sC+\sD, 4-\sA-\sB-\sC);
    \filldraw[estyleweight] (\sA+\sB+\sC+\sD, 4-\sA-\sB-\sC-\sD-\sE) rectangle (\sA+\sB+\sC+\sD+\sE, 4-\sA-\sB-\sC-\sD);
    \filldraw[estyleweight] (\sA+\sB+\sC+\sD+\sE, 0) rectangle (4, \sE);

    \def\barX{4.3} %
    \def\barW{0.7} %
    \filldraw[a1style]  (\barX, 4-\sA) rectangle (\barX+\barW, 4) node[above left, figtextcolor, xshift=-1mm] {\small$L$} node at (\barX+0.5*\barW, 4-0.5*\sA) [black] {\small A1};
    \filldraw[a2style]   (\barX, 4-\sA-\sB) rectangle (\barX+\barW, 4-\sA) node at (\barX+0.5*\barW, 4-\sA-0.5*\sB) [black] {\small A2};
    \filldraw[b1style] (\barX, 4-\sA-\sB-\sC) rectangle (\barX+\barW, 4-\sA-\sB) node at (\barX+0.5*\barW, 4-\sA-\sB-0.5*\sC) [black] {\small B1};
    \filldraw[b2style] (\barX, 4-\sA-\sB-\sC-\sD) rectangle (\barX+\barW, 4-\sA-\sB-\sC) node at (\barX+0.5*\barW, 4-\sA-\sB-\sC-0.5*\sD) [black] {\small B2};
    \filldraw[estyle] (\barX, 1.5*\sE) rectangle (\barX+\barW, 2*\sE) node at (\barX+0.5*\barW, 1.75*\sE) [black, scale=0.9] {\small E11};
    \filldraw[estyle] (\barX, \sE) rectangle (\barX+\barW, 1.5*\sE) node at (\barX+0.5*\barW, 1.25*\sE) [black, scale=0.9] {\small E21};
    \filldraw[estyle] (\barX, 0.5*\sE) rectangle (\barX+\barW, \sE) node at (\barX+0.5*\barW, 0.75*\sE) [black, scale=0.9] {\small E12};
    \filldraw[estyle] (\barX, 0) rectangle (\barX+\barW, 0.5*\sE) node at (\barX+0.5*\barW, 0.25*\sE) [black, scale=0.9] {\small E22};
    
    \draw[{Stealth[length=3mm]}-{Stealth[length=3mm]}, thick, figtextcolor] (2.5, 1.15) .. controls (2.5, 0.5) .. (3.15, 0.5) node[left, figtextcolor, xshift=-4.5mm, yshift=-1.5mm] {\small Weight sharing};
\end{scope}

\draw[-{Stealth[length=3mm]}, thick, figtextcolor] (13.7, 2) -- (15.3, 2) node[midway, above, yshift=1mm] {\small Implementation} node[midway, below, yshift=-1mm] {\small $5.33\times$ fewer} node[midway, below, yshift=-4.5mm] {\small FLOPs};

\begin{scope}[xshift=16cm]
    
    \def\sA{0.5} %
    \def\sB{0.5} %
    \def\sC{0.5} %
    \def\sD{0.5} %
    \def\sE{1.0} %
    \def\fudge{0.25}
    
    \filldraw[a1styleweight] (0, 4-\sA) rectangle (\sA, 4) node at (0, 4-0.5*\sA) [left, figtextcolor] {$\frac{C}{8}$};
    \filldraw[a2styleweight] (0, 4-\fudge-\sA-\sB) rectangle (\sB, 4-\fudge-\sA);
    \filldraw[b1styleweight] (0, 4-2*\fudge-\sA-\sB-\sC) rectangle (\sC, 4-2*\fudge-\sA-\sB) ;
    \filldraw[b2styleweight] (0, 4-3*\fudge-\sA-\sB-\sC-\sD) rectangle (\sD, 4-3*\fudge-\sA-\sB-\sC);
    \filldraw[estyleweight] (0, 4-4*\fudge-\sA-\sB-\sC-\sD-\sE) rectangle (\sE, 4-4*\fudge-\sA-\sB-\sC-\sD) node at (0, 4-4*\fudge-\sA-\sB-\sC-\sD-0.5*\sE) [left, figtextcolor] {$\frac{C}{4}$};

    \def\spacetovec{0.2}
    \def\barW{0.7}
    \filldraw[a1style]  (\sA + \spacetovec, 4-\sA) rectangle (\sA+\spacetovec+\barW, 4) node[above left, figtextcolor, xshift=-1mm] {\small$L$} node at (\sA+\spacetovec+0.5*\barW, 4-0.5*\sA) [black] {\small A1};
    \filldraw[a2style]  (\sB + \spacetovec, 4-\fudge-\sA-\sB) rectangle (\sB+\spacetovec+\barW, 4-\fudge-\sA) node at (\sB+\spacetovec+0.5*\barW, 4-\fudge-\sA-0.5*\sB) [black] {\small A2};
    \filldraw[b1style]  (\sC + \spacetovec, 4-2*\fudge-\sA-\sB-\sC) rectangle (\sC+\spacetovec+\barW, 4-2*\fudge-\sA-\sB) node at (\sC+\spacetovec+0.5*\barW, 4-2*\fudge-\sA-\sB-0.5*\sC) [black] {\small B1};
    \filldraw[b2style]  (\sD + \spacetovec, 4-3*\fudge-\sA-\sB-\sC-\sD) rectangle (\sD+\spacetovec+\barW, 4-3*\fudge-\sA-\sB-\sC) node at (\sD+\spacetovec+0.5*\barW, 4-3*\fudge-\sA-\sB-\sC-0.5*\sD) [black] {\small B2};
    \filldraw[estyle] (\sE + \spacetovec, 4-4*\fudge-\sA-\sB-\sC-\sD-0.5*\sE) rectangle (\sE+\spacetovec+\barW, 4-4*\fudge-\sA-\sB-\sC-\sD) node at (\sE + \spacetovec+0.5*\barW, 4-4*\fudge-\sA-\sB-\sC-\sD-0.25*\sE) [black, scale=0.9] {\small E11};
    \filldraw[estyle] (\sE + \spacetovec, 4-4*\fudge-\sA-\sB-\sC-\sD-\sE) rectangle (\sE+\spacetovec+\barW, 4-4*\fudge-\sA-\sB-\sC-\sD-0.5*\sE) node at (\sE + \spacetovec+0.5*\barW, 4-4*\fudge-\sA-\sB-\sC-\sD-0.75*\sE) [black, scale=0.9] {\small E21};
    \filldraw[estyle] (\sE + \spacetovec + \barW, 4-4*\fudge-\sA-\sB-\sC-\sD-0.5*\sE) rectangle (\sE+\spacetovec+2*\barW, 4-4*\fudge-\sA-\sB-\sC-\sD) node[above left, figtextcolor, xshift=-3.5mm] {\small$2L$} node at (\sE + \spacetovec+1.5*\barW, 4-4*\fudge-\sA-\sB-\sC-\sD-0.25*\sE) [black, scale=0.9] {\small E12};
    \filldraw[estyle] (\sE + \spacetovec + \barW, 4-4*\fudge-\sA-\sB-\sC-\sD-\sE) rectangle (\sE+\spacetovec+2*\barW, 4-4*\fudge-\sA-\sB-\sC-\sD-0.5*\sE) node at (\sE + \spacetovec+1.5*\barW, 4-4*\fudge-\sA-\sB-\sC-\sD-0.75*\sE) [black, scale=0.9] {\small E22};
\end{scope}

\end{tikzpicture}
        }
    }
    \caption{
    \textbf{\octicg Linear layers.}
    Implementing equivariant linear layers in the Fourier domain of $\mathrm{D}_8$ gives a major computational benefit.
    \textbf{Left:} A $C\times C$ weight matrix being multiplied by $L$ tokens of feature dimension $C$.
    \textbf{Center:} The block-diagonalization that happens when enforcing the layer to be $\mathrm{D}_8$-equivariant in the Fourier domain.
    More precisely, we enforce equivariance with respect to the representation $\frac{C}{8}\rho_\text{iso}$ that splits into irreps $\rho_\text{A1}, \rho_\text{A2}, \rho_\text{B1}, \rho_\text{B2}$ and $\rho_\text{E}$ as detailed in Section~\ref{sec:method}.
    There is no mixing between different irreps and the weight sharing in the block-diagonal stems from the fact that $\rho_\text{E}$ is a two-dimensional irrep.
    \textbf{Right:} An efficient implementation of the original $C\times C$ by $C \times L$ matrix multiplication as four $\frac{C}{8}\times\frac{C}{8}$ by $\frac{C}{8}\times L$ and one $\frac{C}{4}\times\frac{C}{4}$ by $\frac{C}{4}\times 2L$ matrix multiplication.
    An equivariant linear layer of this type requires $16/3\approx 5.33$ times fewer FLOPs to compute and has $8$ times fewer parameters than the ordinary linear layer shown to the left.
    }
    \label{fig:linear_layer}
\end{figure}
Equivariant networks use equivariant linear layers, which map between representations.
    A linear $G$-equivariant map or intertwiner, $W\in \mathbb{R}^{n\times n}$, between representations $\rho_1, \rho_2$, commutes with their action, i.e., $\rho_2(g)W = W\rho_1(g)$.
    It follows from Schur's lemma
    that $W = \lambda I$ for some scalar $\lambda$ if $\rho_1, \rho_2$ are irreps, and that $\lambda = 0$ if $\rho_1, \rho_2$ are not isomorphic~\cite[Section~I.2.2]{serre_linear_1977}\footnote{
    We can apply Schur's lemma for complex representations here since the irreps listed in Example~\ref{ex:irreps} are irreducible over the complex numbers.
    We will however only use real-valued linear layers, i.e. $\lambda\in\mathbb{R}$.}.
As $\rho_{\text{iso}}$ is just a stack of irreducible representations, any intertwiner between such representations will be sparse, which is why they are less computationally expensive than ordinary linear layers.
    The naive computational complexity for a linear map $W:\mathbb{R}^{|\octicg|} \to \mathbb{R}^{|\octicg|}$ is $|\octicg|^2=64$ multiplications\footnote{We ignore the additions for simplicity.}. In contrast, intertwiners $\rho_{\text{iso}} \to \rho_{\text{iso}}$ require a total of $\sum_i^k m_i^2d_i = 1+1+1+1+2^2\cdot2 = 12$ multiplications for~\octicg. From a signal processing perspective, this is analogous to convolution being point-wise multiplications in frequency space.
    We visualize the computational savings obtained by working in the Fourier domain of $\mathrm{D}_8$ in Figure~\ref{fig:linear_layer}.

\begin{example}[Images]
    Square images can be considered as elements of $\mathbb{R}^{3 \times M\times M}$ where $3$ is the number of color channels and $M$ is the image height/width in pixels.
    There is a natural \octicg-representation $\rho_\text{image}$ associated with square images, where $\rho_\text{image}(r)$ is the pixel permutation rotating the images anti-clockwise by $\ang{90}$ and $\rho_\text{image}(s)$ is the permutation reflecting the images left-to-right.
\end{example}

\begin{example}[ViT features]
    In ViTs, features are elements of $\mathbb{R}^{C\times N\times N}$, which we will think of as $C\times N^2$ matrices.
    $N$ is the number of image tokens along the height/width of the image, so $N = M/P$ where $P$ is the patch size (typically $P=14$ or $P=16$) and
    $C$ is the channel dimension.
    The simplest representation that we consider on features is the permutation representation $\rho_\text{token}$ that, analogously to $\rho_\text{image}$, permutes the tokens according to elements of \octicg.
\end{example}

\begin{example}[Steerable ViT features]
    We can equip the channel dimension of ViT features with a group representation $\rho_\text{chan}$ to obtain ``steerable'' features.
    If $C$ is divisible by $8$, we can consider multiples of $\rho_\text{reg}$ or $\rho_\text{iso}$ as $\rho_\text{chan}$.
    The complete representation $\rho$ acting on the $C \times N^2$-matrix $\mathbf{x}$, is then permuting the tokens according to $\rho_\text{token}$ and modifying the channels according to $\rho_\text{chan}$.
    Concretely,
    \begin{equation}\label{eq:kronecker_rep}
        \rho(g)\mathrm{Vec}(\mathbf{x}) = \mathrm{Vec}\left(\rho_\text{chan}(g) \mathbf{x} \rho_\text{token}(g)^\mathsf{T}\right)
        =\left(\rho_\text{token}(g) \otimes \rho_\text{chan}(g)\right)\mathrm{Vec}(\mathbf{x}),
    \end{equation}
    where $\mathrm{Vec}(\mathbf{x})$ is the column-wise vectorization of the matrix $\mathbf{x}$
    and $\otimes$ is the tensor product of representations or (equivalently) the Kronecker product of matrices.
    We refer to features transforming according to \eqref{eq:kronecker_rep} as $\rho_\text{chan}$-steerable, or features of type $\rho_\text{chan}$.
    This is a simpler form of the induced representations typically considered in steerable CNNs~\cite{cohen-steerable,weiler_cesa_2019}, the simplification coming from the fact that we don't enforce translation equivariance.
    Steerable ViT-features are illustrated in the Appendix, Figure~\ref{fig:visualizations}.
\end{example}

\begin{example}[Patchification of images] \label{ex:patchification}
    One can consider a patchified image as a steerable ViT feature in the following way.
    By patchification we mean the operation of reshaping a $3 \times M\times M$ image first to $N^2$ patches of size $P\times P$, with $NP=M$ and then to a $3P^2\times N^2$ matrix.
    When we transform the original image by $\rho_\text{image}$, the patchified image is rotated by $\rho_\text{token}$ and $\rho_\text{chan}$ as in \eqref{eq:kronecker_rep}.
    Now $\rho_\text{chan}(g)$ is a permutation matrix that rotates or mirrors a patch and we will denote this particular $\rho_\text{chan}$ by $\rho_\text{patch}$.
\end{example}

\subsubsection{Notation.}
\label{subsec:notation} For convenience of the reader, we collect the most important notation in this section.
We use the bold letter $\mathbf{x}$ for ViT features, which have shape $C\times L$, where $L$ is the number of tokens and $C$ the channel dimension.
Typically, $L=N^2$, where $N$ is the image height/width in tokens, or $L=N^2+1$ with a class token.
For an individual $C$-dimensional token we use the letter $x$ which
is often acted on by $\rho_\text{chan}=\frac{C}{8}\rho_\text{iso}$, in which case we can split $x$ into $C/8$-dimensional sub-tokens $x_\text{A1}, x_\text{A2}, x_\text{B1}, x_\text{B2}, x_\text{E11}, x_\text{E12}, x_\text{E21}, x_\text{E22}$, where the first four transform according to the irreps $\rho_\text{A1}, \rho_\text{A2}, \rho_\text{B1}, \rho_\text{B2}$ respectively while the $2\times \frac{C}{8}$-matrices $(x_\text{E11}\quad x_\text{E12})^\mathsf{T}$ and $(x_\text{E21}\quad x_\text{E22})^\mathsf{T}$ both transform according to $\rho_\text{E}$.

\subsection{Quick ViTs} \label{subsec:octic-vits}

We construct ViT versions that map images to steerable ViT features $\mathbf{x}\in \mathbb{R}^{C\times L}$ in the first layer (PatchEmbed) and then process steerable ViT features in subsequent layers.
We choose $\rho_\text{chan}$ to be a multiple of $\rho_\text{iso}$, enabling efficient linear layers.
Typically, we write $\rho_\text{chan}=\frac{C}{8}\rho_\text{iso}$ where $C$ is the embedding dimension of the ViT.
For classification tasks, we map the steerable ViT features to \octicg invariant features fed into a classification head.
We will also consider networks that use $\rho_\text{chan}$-steerable features for the first layers and then map either to $D\rho_\text{A1}$-steerable features, these networks are denoted $\mathcal{I}_8(\text{ViT})$, or break equivariance, denoted $\mathcal{H}_8(\text{ViT})$ (here $\mathcal{I}$ is short for invariant and $\mathcal{H}$ for hybrid).
We refer to ViTs that fall into these three families broadly as \textit{Quick ViTs} or, as they in our current implementation are specialized for the octic group, \textit{octic ViTs}.
In Section~\ref{sec:equi-inv-hybrid}, we study the effect of varying the number of $\rho_\text{chan}$-equivariant layers.

A commonly appreciated fact is that a transformer component $b$ such as an MLP or Attention layer, is permutation equivariant over tokens, which implies %
\begin{equation}
    b(\mathbf{x}\rho_\text{token}(g)^\mathsf{T}) = b(\mathbf{x})\rho_\text{token}(g)^\mathsf{T}.
\end{equation}
For $b$ to be fully equivariant it also needs to be so in the channel dimension \begin{equation}\label{eq:equiv_twoside}
    b\left(\rho_\text{chan}(g)\mathbf{x}\rho_\text{token}(g)^\mathsf{T}\right) = \rho_\text{chan}(g)b(\mathbf{x})\rho_\text{token}(g)^\mathsf{T}.
\end{equation}
Designing the components of ViT blocks so that \eqref{eq:equiv_twoside} holds is the topic of Section~\ref{subsec:octic-layers}.

\begin{figure}[t]
\vskip 0.2in
\begin{center}
\centerline{\includegraphics[width=1.0\columnwidth]{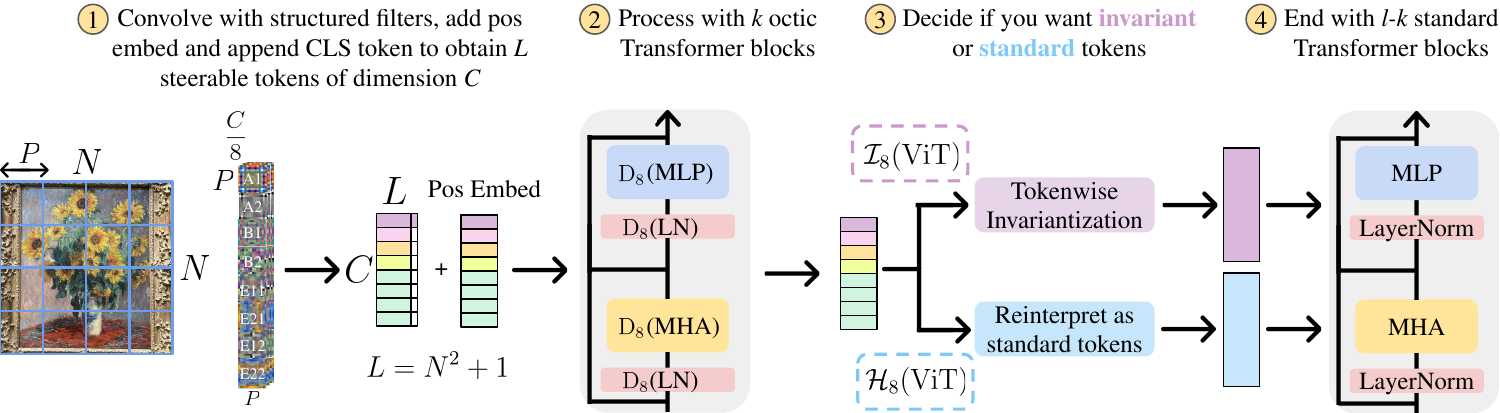}}
\caption{
\textbf{Quick ViT architecture}. Patches are first extracted from an image using specialized octic equivariant filters and the resulting features are processed by $k$ octic equivariant ViT blocks. The final  embeddings can be fed to $l-k$ standard transformer blocks (as demonstrated by our $\mathcal{H}_8$ and $\mathcal{I}_8$ ViTs).
}
\label{fig:arch}
\end{center}
\vskip -0.2in
\end{figure}

\subsection{Octic Equivariant Layers}
\label{subsec:octic-layers}

In this section we detail our implementations of \octicg equivariant transformer layers. Together, these pieces can be combined into a ViT block $b=b_2\circ b_1$ where
\begin{equation}
b_1(\mathbf x) = \mathbf x + \mathrm{MHA}(\mathrm{LN}(\mathbf x)), \quad
b_2(\mathbf x) = \mathbf x + \mathrm{MLP}(\mathrm{LN}(\mathbf x)).
\end{equation}
LN is layer normalization, MHA multi-head self-attention and MLP a $C \to 4C$ linear map, GELU, and $4C \to C$ linear map. The blocks can subsequently be stacked, as illustrated in Figure~\ref{fig:arch}.

\subsubsection{The Patch Embedding Layer.}
The first layer in a ViT, following \cite{dosovitskiy2021an}, is the patch embedding, short PatchEmbed.
In our case, it can be viewed as a mapping from steerable features of type $\rho_\text{patch}$ to steerable features of type $\frac{C}{8}\rho_\text{iso}$ where $C$ is the embedding dimension of the ViT.
It is implemented as a convolution over the input image with kernel size and stride equal to the patch size $P$.
The convolution kernels are weight sharing constrained %
to map to features of the different irreps-types in $\frac{C}{8}\rho_\text{iso}$, see Appendices~\ref{appendix:visualizations} and \ref{appendix:learned-filters} for kernel visualizations.

Directly after PatchEmbed, we add a learnable positional encoding $\mathbf{e}\in \mathbb{R}^{C \times L}$ to the features.
The positional encoding is not constant over tokens, thereby breaking translation equivariance.
To be \octicg equivariant $\mathbf{e}$ must satisfy
\begin{equation}
\begin{split}
    \forall\mathbf{x}:\rho_\text{chan}(g) \mathbf{x} \rho_\text{token}(g)^\mathsf{T} + \mathbf{e}
    &= \rho_\text{chan}(g) (\mathbf{x} + \mathbf{e}) \rho_\text{token}(g)^\mathsf{T} \\
    \Longleftrightarrow \quad 
    \mathbf{e} &= \rho_\text{chan}(g) \mathbf{e}\rho_\text{token}(g)^\mathsf{T}.
\end{split}
\end{equation}
In words, the positional encoding at a specific token position $p$ must be a $\rho_\text{chan}(g)$-transformed version of the positional encoding at the token position that is permuted to $p$ by $\rho_\text{token}(g)$. After adding the positional encoding, we append a learnable class token \texttt{[CLS]}$\in \mathbb{R}^{C \times 1}$ to the features.
To ensure equivariance, we enforce it to be non-zero only in the A1 feature type. 

\subsubsection{Linear Layers.}
Linear layers appear in ViTs both in the MLP block and the MHA block.
As mentioned in Section~\ref{subsec:octic-equi} and illustrated in Figure~\ref{fig:linear_layer}, equivariant linear layers map between irreps of the same type due to Schur's lemma.
This fact was used to construct efficient reflection-equivariant neural networks by \cite{flopping-for-flops}.
Here, we use the same approach for octic equivariance.

To re-iterate, for features of type $\rho_\text{chan}=\frac{C}{8}\rho_\text{iso}$ we consider each token $x\in\mathbb{R}^C$ split into $x_\text{A1}, x_\text{A2}, x_\text{B1}, x_\text{B2}, x_\text{E}$ of dimensions $C/8, C/8, C/8, C/8$ and $2\times C/4$.
Linear layers map each irrep type to itself, meaning that they are parameterised by four $C/8\times C/8$ matrices and one $C/4\times C/4$ matrix, yielding a factor $8$ fewer parameters than a general linear layer.

In terms of FLOPs needed to compute the linear layer, $x_i \mapsto W_i x_i$ requires $C/8\cdot C/8=C^2/64$ FLOPs for $i\in\{\text{A1, A2, B1, B2}\}$ while due to the ``weight-sharing'' over the two dimensions in $\rho_\text{E}$, it requires $2\cdot C/4\cdot C/4=C^2/8$ FLOPs for $i=\text{E}$.
In total we therefore get $16/3\approx 5.33$ times fewer FLOPs than the $C^2$ required for an ordinary linear layer.

\subsubsection{Activation Functions, Layer Norm, Attention, and Invariantization.}
A pointwise activation function $\sigma$ can be applied equivariantly after transforming the features from the Fourier domain (multiples of $\rho_\text{iso}$), to the spatial domain (multiples of $\rho_\text{reg}$), as discussed in Example~\ref{ex:isotypic}.
In the spatial domain $\sigma$ can be applied point-wise as this commutes with the permutation representation $\rho_\text{reg}$.
In Quick ViTs, following prior work, we apply the GELU activation function. 

We implement (token-wise) equivariant layer normalization by transforming each irrep separately to mean $0$ followed by division of %
norm over all the irreps.

If two tokens $q$ and $k$ transform according to the same orthogonal representation $\rho_\text{chan}$, then $q^\mathsf{T} k$ is invariant under \octicg since $(\rho_\text{chan}(g)q)^\mathsf{T}(\rho_\text{chan}(g)k) = q^\mathsf{T}\rho_\text{chan}(g)^\mathsf{T}\rho_\text{chan}(g)k=q^\mathsf{T}k$.
This means that the computation of attention logits in ordinary scaled dot-product attention is invariant, so the subsequent weighted sum over value tokens is equivariant.

To output \octicg invariant classification predictions, we map from features of type $\frac{C}{8}\rho_\text{iso}$ to features of type $C\rho_\text{A1}$ and then extract the \texttt{[CLS]} token.
We ablate different invariantizations to A1-tokens in Appendix~\ref{appendix:invariantization}, including linear invariants, triple correlation~\cite{kakarala2012bispectrum,sanborn2023general}, max filtering~\cite{cahill2024group}, generators of the ring of invariant polynomials and canonisation of the signal~\cite{kaba_equivariance_2023}.
We find that a power spectrum invariantization works well and settle on that for the %
the experiments.

\section{Computational Efficiency}
\label{sec:comp-eff}
\begin{figure}[t]
  \centering
  \begin{subfigure}[b]{0.45\linewidth}
    \centering
    \includegraphics[width=\linewidth]{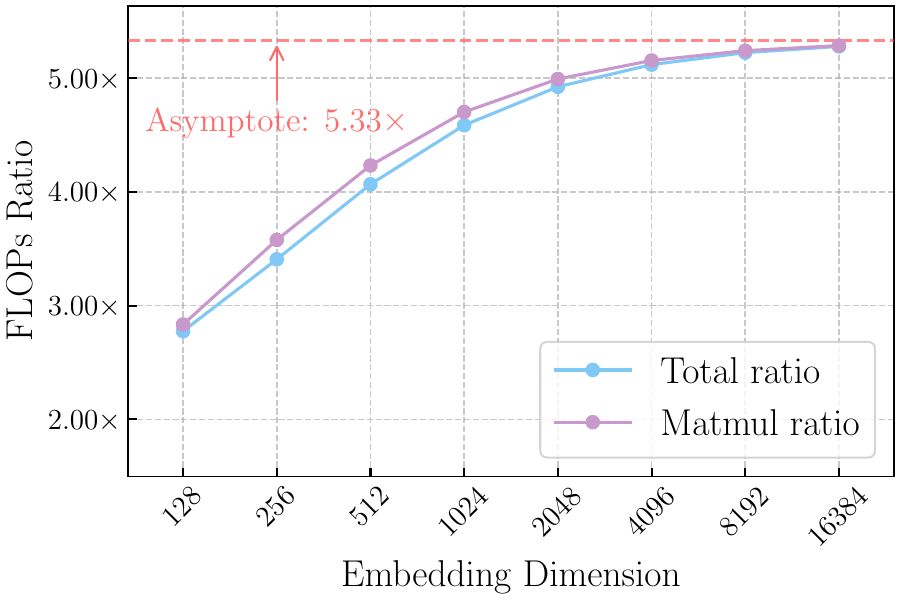}
    \caption{FLOPs ratio vs. embedding dimension}
    \label{fig:flops-ratio}
  \end{subfigure}
  \begin{subfigure}[b]{0.45\linewidth}
    \centering
    \includegraphics[width=\linewidth]{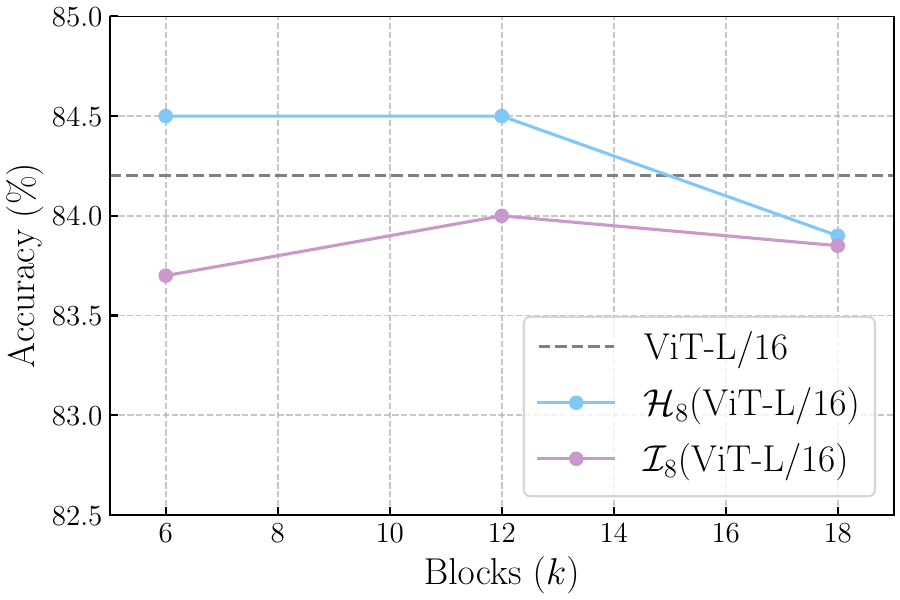}
    \caption{Accuracy vs. number of octic blocks ($k$)}
    \label{fig:hybridisation_ablation}
    
  \end{subfigure}
  \caption{
    {\textbf{(a)} Reduction in FLOPs from a non-equivariant transformer block to an octic-equivariant block vs.\ embedding dimension. The matmul ratio reflects only matrix multiplications in linear layers and Attention; the total ratio includes all computations. \textbf{(b)} The effect of changing number of octic blocks ($k$) for ViT-L, out of $l=24$ blocks.}
  }
\end{figure}

As transformers scale, the linear layers dominate the execution time~\cite{kaplan2020scaling}. Thus, as ViTs grow, the FLOPs savings will approach those of the linear layer, a reduction of $5.33$ times.
We plot the FLOPs savings of a ViT block as the embedding dimension increases in Figure~\ref{fig:flops-ratio}.
The computational benefits of equivariant ViTs are more pronounced at scale, and we benchmark the throughput of various ViT sizes from the literature and their equivariant counterparts in Table \ref{tab:vit-scale}. 
In this throughput benchmark we replace all of the ViT blocks with equivariant blocks, and therefore the models fall into the $\mathcal{I}_8$ family of networks, but with invariantization after the last block rather than half the blocks.

\begin{table}[t]
\centering
     \caption{\textbf{Compute scaling.} We measure the scaling of \octicg-equivariant ViTs. The model sizes are taken from \cite{scalingvits22b:2023} and we do not train the largest models as part of this work. The numbers ending in ``x'' describe the change from standard ViT statistics of the corresponding equivariant ViTs.
     We benchmark in half precision.
     In full precision speed gains are much more pronounced, especially at small scales, c.f.\ Table~\ref{tab:linear-performance}.
     In this table we show models where all linear layers are replaced by block-diagonal layers.
     In Section~\ref{sec:experiments} we train models with only half of the layers block-diagonal.
     Hence the throughput improvements there are lower.
     }
  \centering
  \begin{tabular}{@{}lcccccccc}
    \toprule 
    Model & Width & Depth & MLP-dim & Heads & Speed(bf16) & FLOPs & Peak Mem. \\
    \midrule	
    ViT-L & 1024 & 24 & 4096 & 16 & 1.32x & 1/4.58x & 1/2.44x \\
    ViT-H & 1280 & 32 & 5120 & 16 & 1.47x & 1/4.58x & 1/2.88x \\
    ViT-G & 1664 & 48 & 8192 & 16 & 1.91x & 1/4.88x & 1/4.36x  \\
    ViT-e & 1792 & 56 & 15360 & 16 & 2.37x & 1/5.01x & 1/5.30x  \\
    ViT-22B & 6144 & 36 & 24576 & 48 & 3.54x & 1/5.18x & 1/5.80x  \\
    \bottomrule		
  \end{tabular}
    \label{tab:vit-scale}
\end{table}

As shown in Table~\ref{tab:vit-scale}, savings in FLOPs do not translate one-to-one to improvements in throughput (images per second) in our current implementation.
However, it is still the case that the throughput is greatly improved.
Our Quick ViT models are pure PyTorch with \texttt{torch.compile}, except for the GELU nonlinearity.
We implement a custom Triton~\cite{triton:2019} kernel that fuses the GELU nonlinearity with the Fourier and inverse Fourier transforms, limiting memory transfers and kernel invocation overhead. While ordinary GELU is pointwise, the new fused Triton kernel is eight to eight points. For extra efficiency, we implement the Fourier transforms by a FFT on \octicg, described in Appendix~\ref{appendix:fourier-transform}.
In Appendix~\ref{appendix:timing-gelu} we show that the overhead of our nonlinearity compared to standard GELU is minimal when compared to the computational complexity of the linear layers.
The gap between savings in FLOPs and speedup is in fact due to the linear layers not having sufficient arithmetic complexity at low channel dimensionality as demonstrated in the next section.

\subsection{Arithmetic Intensity}\label{appendix:arithmetic-intensity}
The arithmetic intensity of an operation is the FLOPs per transferred byte to the GPU and it can be compared with the FLOPs per bandwidth of a given device to obtain a bound on the maximum achievable throughput~\cite{scaling-book}.
The arithmetic intensities are $\frac{2BCF}{P(BC+CF+BF)}$ and $\frac{2BCF/5.33}{P(BC+CF/8+BF)}$ for the standard and octic linear layers, respectively, where $B$ is the batch size in tokens, $C$ is the input dimension, $F$ is the output dimension and $P$ is the precision in bytes.
This means that the octic and ordinary layers scale differently.
At large scale, not only FLOPs are improved by octic layers but also arithmetic intensity.
For instance, for $B=196$ (one image worth of tokens), $P=2$ (half precision) and $F=4C$ (a typical MLP expansion factor) one can calculate that ordinary linear layers have higher arithmetic intensity up to $C\approx 3200$, whereas octic linear layers have higher arithmetic intensity at thereafter.
For the experiments in this paper, we are not able to scale to such large dimensions, but still get throughput benefits due to savings in FLOPs, as shown in Table~\ref{tab:vit-scale}.

For larger batch sizes, standard linear layers typically have larger arithmetic intensity than octic linear layers, because the terms without $B$ in the denominator become negligible. However, at sufficiently large $C$ both standard and octic linear layers are compute bound and hence what matters for throughput is the total number of FLOPs, making octic linear layers $5.33$ times faster.

\begin{table}[t]
    \centering
     \caption{\textbf{Benchmarking linear layers.} Performance of three different types of linear layers over different precisions and different amount of channels.}
    \label{tab:linear-performance}
  \begin{tabular}{lccrrrr}
        \toprule
        & & & AI~~~~~~~~ & Time & Perf. & \\
        Linear layer & Precision & Channels & \footnotesize{(FLOPs/byte)} & \footnotesize{(ms)} & \footnotesize{(TFLOPs/s)} & Speedup  \\
        \midrule
        Standard & fp16 & $1024 \to 4096$ & 770 & 0.47 & 230 & 1.0 \\
        \octicg & fp16 & $1024 \to 4096$ & 150 & 0.19 & 110 & 2.6 \\
        Eight blocks & fp16 & $1024 \to 4096$ & 100 & 0.13 & 100 & 3.6 \\
        \midrule
        Standard & fp16 & $8192 \to 32768$ & 4300 & 24 & 280 & 1.0 \\
        \octicg & fp16 & $8192 \to 32768$ & 1200 & 5.3 & 240 & 4.6 \\
        Eight blocks & fp16 & $8192 \to 32768$ & 770 & 3.4 & 250 & 7.2 \\
        \midrule
        Standard & fp32 & $1024 \to 4096$ & 380 & 5.6 & 19 & 1.0 \\
        \octicg & fp32 & $1024 \to 4096$ & 80 & 1.2 & 17 & 4.8 \\
        Eight blocks & fp32 & $1024 \to 4096$ & 50 & 0.76 & 17 & 7.4 \\
        \midrule
        Standard & fp32 & $8192 \to 32768$ & 2200 & 360 & 19 & 1.0 \\
        \octicg & fp32 & $8192 \to 32768$ & 580 & 67 & 19 & 5.33 \\
        Eight blocks & fp32 & $8192 \to 32768$ & 380 & 45 & 19 & 7.95 \\
        \bottomrule
    \end{tabular}
\end{table}

As mentioned, the main bottleneck to achieve throughput gains closer to the actual FLOP reductions in octic ViTs is that the linear layers are not performing sufficiently well for low feature dimensions $C$.
To more clearly illustrate the bottleneck and tie it to the arithmetic intensity (AI) of the layers we conduct a simple throughput benchmark for only linear layers.
We include standard linear layers, octic equivariant linear layers and block-diagonal layers with eight equal size blocks, all from $C$ to $F=4C$ channels.
The block-diagonal layers have arithmetic intensity $\frac{2BCF/8}{P(BC+CF/8+BF)}$.

We benchmark with batch size $64\cdot 196$ (i.e. $64$ images worth of tokens) on an A100-80GB GPU.
The performance in bf16 and fp16 precision is equal with an advertised maximum achievable performance of $312$ TFLOPs/s.
We use bf16 precision when training the networks in this paper.
For reference we also include results in fp32 precision, where the advertised maximum achievable performance is $19.5$ TFLOPs/s.
The ridge point for when computations become theoretically compute bound is at $156$ FLOPs/byte for the lower precision and $10$ FLOPs/byte for the higher precision.
We see in Table~\ref{tab:linear-performance} that for non-compute bound layers such as octic and block-diagonal linears at $C=1024$ in fp16, the performance in TFLOPs/s is far below the compute bound layers, leading to a lower speedup than predicted by the FLOP counts. However, in fp32 the speedup is drastically higher.

\section{Experiments} \label{sec:experiments}

In this section, we evaluate our Quick ViTs on supervised (DeiT III) and self-supervised (DINOv2) training recipes and perform ablations. DeiT III~\cite{deitiii:2022} is a popular and highly tuned supervised training recipe for classification and DINOv2~\cite{dinov2} is a state-of-the-art self-supervised method to extract visual features at large scale. All models are trained on ImageNet-1K~\cite{imageNet2009, russakovsky2015imagenet, imageNet2019} following the official implementations. Code is available at \href{https://github.com/davnords/octic-vits}{https://github.com/davnords/octic-vits}. 

\subsection{DeiT III}

We train an array of networks on the supervised task of image classification and compare to the performance reported by \cite{deitiii:2022} and \cite{flopping-for-flops}. We find, as illustrated in Table~\ref{tab:deit}, that incorporating octic-equivariant layers provides significant computational savings without sacrificing accuracy. In particular, our $\mathcal{H}_8(\text{ViT-H/14})$ model achieves a classification performance of 85.0\% compared to the baseline of 84.6\% while using only 61\% of the FLOPs and matching the performance of the $\mathcal{H}_2(\text{ViT-H/14})$ that incorporates flopping ($\mathrm{D}_2$) equivariance while being more computationally efficient. Similar computational gains are obtained with the invariant model $\mathcal{I}_8(\text{ViT-H/14})$, which achieves a performance of 84.7\%.

In the final column of Table~\ref{tab:deit}, we study the effect of evaluating models on a randomly rotated validation set without training on such augmentations. We find that the invariant model performs equally well while the performance of the remaining models (including $\mathcal{H}_8$) significantly degrade.

\begin{table}[t]
\centering
     \caption{
\textbf{DeiT III evaluation.} We measure the Top-1 classification accuracy on ImageNet-1K for different model sizes. Our networks are marked with $\dagger$.
\label{tab:deit}}
  \centering
  \begin{tabular}{@{}lrrrr@{} r rr@{\ }}
    \toprule 
    Model & params & throughput bf16 & FLOPs & Peak Mem. &&
    Top-1 & OOD Rot. \\
     & ($\times 10^6$) & (im/s) & ($\times 10^9$) & (MB) && Acc. $\uparrow$ & $\Delta$Acc.$\uparrow$ \\
    \midrule
    ViT-H/14 & 632.1 & 569 & 167.8 & 3285 && 84.6 & -12.6 \\
    $\mathcal{H}_2(\text{ViT-H/14})$  & 474.2 & 640 & 127.4 & 2734 && \bfseries 85.0 & -13.4 \\
    $\mathcal{I}_8(\text{ViT-H/14})^\dagger$  & 362.3 & 657 & 104.0 & 2249 && 84.7 & \bfseries0.0 \\
    $\mathcal{H}_8(\text{ViT-H/14})^\dagger$  & 355.8 & 660 & 102.3 & 2223 && \bfseries85.0 & -13.4 \\
    \midrule
    ViT-L/16  & 304.4 & 1421 & 61.9 & 1557 && 84.2 & -13.4 \\
    $\mathcal{H}_2(\text{ViT-L/16})$  & 228.3 & 1610 & 46.9 & 1458 && \bfseries84.5 & -13.9 \\
    $\mathcal{I}_8(\text{ViT-L/16})^\dagger$  & 175.5 & 1618 & 38.5 & 1210 && 84.0 & \bfseries0.0 \\
    $\mathcal{H}_8(\text{ViT-L/16})^\dagger$  & 171.3 & 1615 & 37.7 & 1194 && \bfseries84.5 & -13.6 \\
    \bottomrule		
  \end{tabular}
\end{table}

\begin{table}[t]
\centering
     \caption{
\textbf{DINOv2 evaluation.} We evaluate the frozen DINOv2 features by classification accuracy on ImageNet-1K (IN1K) and segmentation mIoU on ADE20K~\cite{ade20k:2017, ade20k:2019} and VOC2012~\cite{pascal-voc-2012}. Our networks are marked with $\dagger$.
\label{tab:dinov2}}
  \centering
    \begin{tabular}{@{}lcccc@{} c cc c cc c cc@{\ }}
        \toprule 
        Model & FLOPs &&
        \multicolumn{2}{@{}c@{}}{IN1K (acc.) $\uparrow$} && \multicolumn{2}{@{}c@{}}{ADE20K (mIoU) $\uparrow$} && \multicolumn{2}{@{}c@{}}{VOC2012 (mIoU) $\uparrow$}\\
        \cmidrule{4-5}
        \cmidrule{7-8}
        \cmidrule{10-11}
         & ($\times 10^9$) && linear & $k$-NN && linear & $k$-NN && linear & $k$-NN \\
        \midrule
        ViT-H/16 & 127.7 && 81.7 & 81.0 && 34.7 & 30.6 && 70.7 & 60.9\\
        $\mathcal{H}_2(\text{ViT-H/16})$ & 97.0 && 81.9 & 80.9 && 34.8 & 30.8 && 70.7 & 61.2\\
        $\mathcal{I}_8(\text{ViT-H/16})^\dagger$ & 77.7 && 81.9 & 80.9 && 33.9 & 29.2 && 70.6 & 61.2 \\
        $\mathcal{H}_8(\text{ViT-H/16})^\dagger$ & 77.5 && \bfseries82.2 & \bfseries81.4 && \bfseries35.1 & \bfseries31.1 && \bfseries70.8 & \bfseries61.7 \\
        \midrule
        ViT-L/16 & 61.9 && 80.9 & 80.5 && 33.2 & 28.4 && 69.1 & 58.1 \\
        $\mathcal{H}_2(\text{ViT-L/16})$ & 46.9 && \bfseries 81.3 & 80.7 && 33.4 & 29.0 && 69.3 & 58.0 \\
        $\mathcal{I}_8(\text{ViT-L/16})^\dagger$  & 38.5 && 81.2 & 80.4 && 32.6 & 28.0 && \bfseries70.0 & \bfseries59.6 \\
        $\mathcal{H}_8(\text{ViT-L/16})^\dagger$ & 37.7 && \bfseries81.3 & \bfseries80.8 && \bfseries33.6 & \bfseries29.4 && 69.1 & 59.5 \\
    
        \bottomrule		
      \end{tabular}
\end{table}

\subsection{DINOv2}

As another pretraining task, we consider the DINOv2 recipe and train our own baselines.\footnote{Note that the pretraining objectives for DINOv2~\cite{dinov2} and DINOv3~\cite{simeoni2025dinov3} are identical.}
Results are summarized in \cref{tab:dinov2}. We find that incorporating octic-equivariant layers maintains or improves downstream classification and segmentation performance while saving FLOPs. In particular, our invariant network $\mathcal{I}_8(\text{ViT-H/16})$ matches the downstream performance of the baseline while using only $61\%$ of the FLOPs and $\mathcal{H}_8(\text{ViT-H/16})$ slightly improves performance with similar savings. 

In Appendix~\ref{appendix:dinobloom}, we further investigate the performance of our invariant model and evaluate it on white blood cell classification, a task that lacks a canonical orientation. The invariant model $\mathcal{I}_8(\text{ViT-L/16})$ outperforms the baseline on most evaluated metrics.

\subsection{Ablations}
\label{sec:equi-inv-hybrid}

\subsubsection{Impact of Equivariance.}%
\label{sec:equi_break}

We study the effect of equivariance by replacing the kernel constrained equivariant patch embed by an arbitrarily linear mapping while keeping the rest of the architecture the same as for $\mathcal{H}_8$. In principle, the model can learn to be equivariant. We evaluate ViT-B using the DeiT III recipe and achieve an accuracy of $82.4$ compared to $83.0$ for $\mathcal{H}_8$(ViT-B). The results suggest that equivariance yields higher accuracy than arbitrary mappings with the same block-diagonal structure (in addition to providing steerable features that can be useful in downstream applications).

\subsubsection{Number of Octic Blocks.}
\label{sec:nbr_octic_blocks}
We experiment with incorporating non-equivariant blocks following~\cite{weiler_cesa_2019}. We include networks where the first $k$ of the ViT blocks are octic and the remaining $l-k$ are standard blocks. Figure \ref{fig:hybridisation_ablation} ablates different values of $k$ for ViT-L ($l=24$). We find that $k=\frac{l}{2}$ strikes a good balance between computational efficiency and representational power and use this throughout the paper. Note that $\mathcal{I}_8$ performs worse for small $k$ due to early invariantization.

We find that the $\mathcal{H}_8$-networks typically outperform $\mathcal{I}_8$-networks. 
This usefulness of symmetry breaking corroborates findings in prior work, \eg~\cite{weiler_cesa_2019, vadgama2025utility}.
The intuition is that ImageNet is not a rotationally invariant dataset, so it leads to improved performance to let the network use the fact that the images are upright as extra information for solving the classification task.

\section{Limitations} \label{sec:limitations}
In this work, we limit our scope to an extensive study of the \octicg group and leave larger dihedral groups for future work. A theoretical discussion regarding scaling to larger dihedral groups is given in Appendix~\ref{appendix:larger-dihedral-groups} for reference.
We follow baseline training recipes without hyperparameter tuning, and we do not conduct an extensive ablation study of the share of features per irrep or scale the size beyond ViT-H.

\section{Conclusion}
We introduced octic-equivariant ViT layers that, when incorporated,  maintain accuracy while significantly reducing computational complexity. We validated our proposed architectures by their effectiveness in both supervised and self-supervised learning, and conducted ablation studies to isolate the effect of invariantization, equivariance, and the number of octic blocks. In particular, we achieved an approximate 40\% reduction in FLOPs for ViT-H without sacrificing accuracy, positioning Quick ViTs as a strong addition to the catalog of vision architectures.

For future work, we intend to incorporate Quick ViTs into a broader range of computer vision applications, in particular in settings where geometric equivariance provides a natural inductive bias. Promising directions include 3D vision tasks, such as multi-view masked image modeling~\cite{nordstrom2026mum} and feed-forward 3D reconstruction\cite{dust3r:2024,vggt:2025}, as well as image matching, where robustness to rotations and other geometric transformations is central~\cite{bokman2022case,bokman2024steerers,nordstrom2026whohandlesorientation}. These directions would further clarify the role of equivariant architectures as efficient and geometry-aware backbones beyond image classification.

\section*{Acknowledgements}
This work was supported by the Wallenberg Artificial
Intelligence, Autonomous Systems and Software Program
(WASP), funded by the Knut and Alice Wallenberg Foundation, and by the strategic research environment ELLIIT, funded by the Swedish government. 
The computational resources were provided by the
National Academic Infrastructure for Supercomputing in
Sweden (NAISS) at C3SE, partially funded by the Swedish Research
Council through grant agreement no.~2022-06725, and by
the Berzelius resource, provided by the Knut and Alice Wallenberg Foundation at the National Supercomputer Centre.

\bibliographystyle{splncs04}
\bibliography{main}

\clearpage
\setcounter{page}{1}
\setcounter{section}{0}

\renewcommand{\thesection}{\Alph{section}}

\begin{center}
{\Large\bf Quick ViTs: Speeding up Vision Transformers through Equivariance}\\[0.5em]
{\large Supplementary Material}
\end{center}

\section{\texorpdfstring{Fourier Transform for \octicg}{Fourier Transform for D8}}
\label{appendix:fourier-transform}

\subsection{Implementation}

The Inverse Fourier Transform, i.e., changing basis from the Isotypical representation $\rho_\text{iso}$ to the regular representation $\rho_\text{reg}$, can be written in the case of \octicg as
\begin{equation}
        Q_\text{reg} = \frac{\sqrt{2}}{4}\begin{pmatrix}
            1 & 1 & 1 & 1 & 1 & 1 & 1 & -1 \\
            1 & 1 & -1 & -1 & 1 & -1 & -1 & -1 \\
            1 & 1 & 1 & 1 & -1 & -1 & -1 & 1 \\
            1 & 1 & -1 & -1 & -1 & 1 & 1 & 1 \\
            1 & -1 & 1 & -1 & -1 & 1 & -1 & -1 \\
            1 & -1 & -1 & 1 & -1 & -1 & 1 & -1 \\
            1 & -1 & 1 & -1 & 1 & -1 & 1 & 1 \\
            1 & -1 & -1 & 1 & 1 & 1 & -1 & 1 \\
        \end{pmatrix}.
    \end{equation}

In practice, we use a fast Triton-compiled implementation of the mapping $x\mapsto Q_\text{reg} x$ as shown in Listing~\ref{lst:q_reg}, and similarly for $x\mapsto Q_\text{reg}^\mathsf{T}x$.

\definecolor{slbackground}{HTML}{FDF6E3}
\definecolor{slbase03}{HTML}{002B36}
\definecolor{slbase02}{HTML}{073642}
\definecolor{slbase01}{HTML}{586E75}
\definecolor{slbase00}{HTML}{657B83}
\definecolor{slbase0}{HTML}{839496}
\definecolor{slbase1}{HTML}{93A1A1}
\definecolor{slbase2}{HTML}{EEE8D5}
\definecolor{slbase3}{HTML}{FDF6E3}
\definecolor{slyellow}{HTML}{B58900}
\definecolor{slorange}{HTML}{CB4B16}
\definecolor{slred}{HTML}{DC322F}
\definecolor{slmagenta}{HTML}{D33682}
\definecolor{slviolet}{HTML}{6C71C4}
\definecolor{slblue}{HTML}{268BD2}
\definecolor{slcyan}{HTML}{2AA198}
\definecolor{slgreen}{HTML}{859900}

\lstset{
  language=Python,
  basicstyle=\ttfamily\scriptsize,
  numbers=left,
  numberstyle=\tiny\color{slbase0},
  stepnumber=1,
  numbersep=5pt,
  backgroundcolor=\color{slbackground},
  showspaces=false,
  showstringspaces=false,
  showtabs=false,
  frame=single,
  breaklines=true,
  breakatwhitespace=true,
  caption={Python implementation of $Q_\text{reg}$.\\},
  tabsize=4,
  commentstyle=\textit\color{slbase0},
  keywordstyle=\bfseries\color{slblue},
  stringstyle=\color{slgreen},
  commentstyle=\color{slbase01}\textit,
  morekeywords={self, True, False, None}
}

\begin{lstlisting}[label=lst:q_reg]
import math
SQRT2_OVER_4 = math.sqrt(2) / 4

def isotypical_to_regular(
    x_A1, x_A2, x_B1, x_B2, x_E11, x_E12, x_E21, x_E22
):
    a = x_A1 + x_A2
    b = x_A1 - x_A2
    c = x_B1 + x_B2
    d = x_B1 - x_B2
    e = x_E11 + x_E12
    f = x_E11 - x_E12
    g = x_E21 + x_E22
    h = x_E21 - x_E22
    apc = a + c
    amc = a - c
    bpd = b + d
    bmd = b - d
    eph = e + h
    emh = e - h
    fpg = f + g
    fmg = f - g
    return (
        SQRT2_OVER_4 * (apc + eph),
        SQRT2_OVER_4 * (amc + fmg),
        SQRT2_OVER_4 * (apc - eph),
        SQRT2_OVER_4 * (amc - fmg),
        SQRT2_OVER_4 * (bpd - fpg),
        SQRT2_OVER_4 * (bmd - emh),
        SQRT2_OVER_4 * (bpd + fpg),
        SQRT2_OVER_4 * (bmd + emh)
    )
\end{lstlisting}

\subsection{Time complexity}
\label{appendix:timing-gelu}

The time complexity of FFT/iFFT is linear with respect to $C$ and log-linear with respect to the order of the dihedral group, but since the group remains constant in this work, we focus the study on the time complexity with respect to the embedding dimension $C$. 
The complexity of iFFT$\to$GELU$\to$FFT is thus linear in $C$, which is the same as the complexity of just using GELU.
In contrast, the complexity of linear layers is quadratic in $C$, so savings in the linear layers will more than compensate additional computation in the nonlinearity when $C$ grows.

We benchmark actual runtime on an A100 GPU. We compare the two non-linearities and full MLP blocks (Linear($C$, $4C$) - Nonlinearity - Linear($4C$, $C$)), for various $C$. Times are in $\mu$s, averaged over 1000 runs of a forward pass with batch size 32. Non-linearities are run on embedding dimension $4C$. We report the results in Table~\ref{tab:non-linearity-runtime}.

Consistent with previous results, we find that the equivariant linear layers pay off more with increasing embedding dimension. The extra computations needed for the non-linearity give only a few percent overhead while the savings from the linear layers provide substantial performance gains.

\begin{table}[t]
\centering
     \caption{\textbf{Time complexity of non-linearity.} Comparing the time (in $\mu$s) of GELU and the MLP block for standard and Quick ViT implementations. The extra computations needed for the non-linearity are noticed at small scale but as $C$ grows the savings from the linear layers dominate.}
  \centering
  \begin{tabular}{@{}lccccc}
    \toprule 
    Model & $C=256$ & $C=512$ & $C=1024$ & $C=2048$ \\
    \midrule	
    GELU & 44.8 & 82.8 & 155.8 & 302.6 \\
    GELU$_{\octicg}$ & 46.1 & 84.7 & 158.8 & 307.0 \\
    MLP & 128.4 & 324.2 & 1053.3 & 3644.9 \\
    MLP$_{\octicg}$ & 142.4 & 244.4 & 471.4 & 1123.4 \\
    \bottomrule		
  \end{tabular}
    \label{tab:non-linearity-runtime}
\end{table}

\section{Training Speed Comparisons}
\label{appendix:training-speed}

In the main manuscript only inference speed is measured (e.g. see Tables~\ref{tab:deit} and \ref{tab:vit-scale}). In Table~\ref{tab:training-speed}, we show that inference throughput gains also translate into training (forward and backward) improvements by measuring the training throughput (images/second). For fair comparison we showcase mixed precision training as this is the modern standard. However, in full precision, the throughput gains are significantly greater.

\begin{table}[t]
    \centering
    \caption{\textbf{Training speed}. We measure the training speed (forward + backward) when using Quick ViTs in different training pipelines. We report the throughput (images/second) and the speed gain compared to the baseline. We run on an A100-80GB in mixed precision.}
    \label{tab:training-speed}
    \begin{tabular}{l rr r rr}
        \toprule
        Model & \multicolumn{2}{@{}c@{}}{DeiT III} && \multicolumn{2}{@{}c@{}}{DINOv2} \\
        \cmidrule{2-3}
        \cmidrule{5-6} & Throughput & Gain && Throughput & Gain \\
        \midrule
        $\text{ViT-H}$                & 180 & $1.00\times$ && 94 & $1.00\times$  \\
        $\mathcal{I}_8(\text{ViT-H})$ & 207 & $1.15\times$ && 114 & $1.21\times$ \\
        $\mathcal{H}_8(\text{ViT-H})$ & 209 & $1.16\times$ && 115 & $1.22\times$ \\
        $\mathrm{D}_8(\text{ViT-H})$  & 257 & $1.43\times$ && 142 & $1.51\times$ \\
        \midrule
        $\text{ViT-L}$                & 454 & $1.00\times$ && 184 & $1.00\times$ \\
        $\mathcal{I}_8(\text{ViT-L})$ & 510 & $1.12\times$ && 210 & $1.14\times$ \\
        $\mathcal{H}_8(\text{ViT-L})$ & 512 & $1.13\times$ &&  214 & $1.16\times$\\
        $\mathrm{D}_8(\text{ViT-L})$  & 564 & $1.24\times$ && 232 & $1.26\times$ \\
        \bottomrule
    \end{tabular}
\end{table}

\section{Larger Dihedral Groups}
\label{appendix:larger-dihedral-groups}
In this section, we briefly sketch the generalization from the octic group to larger dihedral groups.
We will consider the dihedral group $\mathrm{D}_{4n}$ of order $4n$, meaning $180/n$ degree rotations and reflections.
This group has $4$ one dimensional irreps and $n-1$ two-dimensional irreps.
The theoretical discussion from Section~\ref{sec:method} carries over almost as is.
The real irreps are all realizable over the complex numbers, so Schur's lemma implies that linear layers from $\frac{C}{4n}$ copies of the regular representation to itself have $4$ blocks of size $\frac{C}{4n}$ and $n-1$ twice repeated blocks of size $\frac{C}{2n}$.
In Figure~\ref{fig:linear_layer} we see the case $n=2$ for the octic group.

The number of multiplications required for an equivariant linear layer is then $4\cdot(\frac{C}{4n})^2+2\cdot(n-1)\cdot(\frac{C}{2n})^2=\frac{C^2(2n-1)}{4n^2}$.
The compute savings relative to an ordinary linear layer are therefore $\frac{4n^2}{2n-1}$ times fewer FLOPs.

The arithmetic intensity for an equivariant linear layer from $C$ to $F$ channels is $\frac{2BCF(2n-1)/(4n^2)}{P(BC + CF/(4n) + BF)}$.
Just as discussed for the octic group in Section~\ref{appendix:arithmetic-intensity}, whether the full FLOP saving of the linear layers can be realized or not depends on $B, C, F, P$ and the specific hardware used.

\section{Visualizations}
\label{appendix:visualizations}
\begin{figure}[t]
\vskip 0.2in
  \begin{minipage}[b]{0.38\textwidth}
    \centering
    \includegraphics[width=.85\columnwidth]{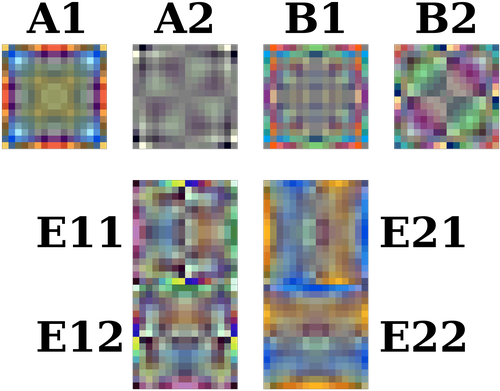}
    \\
    \vspace{0.5em}
    {(a) \textbf{PatchEmbed filters.}} %
    \vspace{0.5em}
    \caption{(a) PatchEmbed filters from a trained network. More filters are shown in Figure~\ref{fig:filters-comparison}.
    \\
    (b-c) Cayley diagrams showing the action of \octicg on (b) patchified images and (c) $\rho_\text{iso}$-features.
    Blue arrows mean horizontal mirroring, $s$, while orange arrows mean mirroring in the bottom-left to top-right diagonal, $sr$.
    The features were obtained by applying the filters in (a) to the patches in (b).}
    \label{fig:visualizations}
  \end{minipage}%
  \hfill%
  \begin{minipage}[b]{0.58\textwidth}
    \centering
    \includegraphics[width=\columnwidth]{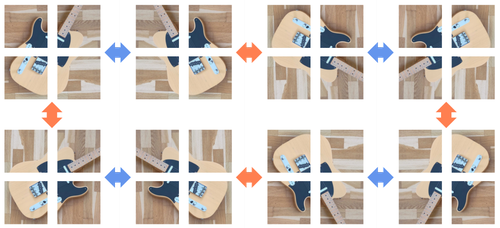}
    \\
    {(b) \textbf{Octic action on patchified images.}} %
    \\
    \vspace{0.8em}
\includegraphics[width=\columnwidth]{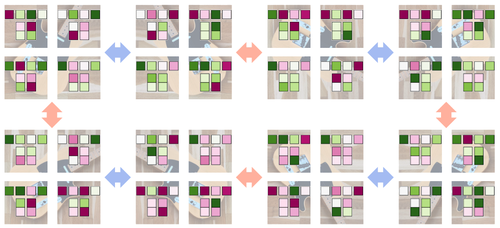}
    {(c) \textbf{Octic action on} $\rho_\text{iso}$\textbf{-features.}} %
    \vspace{1em}
  \end{minipage}
\end{figure}

We visualize the action of the octic group on images and on $\rho_\text{iso}$-features in Figure~\ref{fig:visualizations}.
\subsection{Learned Filters} \label{appendix:learned-filters}

\begin{figure}[htbp]
  \centering

  \caption*{\textbf{(a) Quick ViT DeiT III Learned Filters}}

  \begin{subfigure}[b]{0.25\linewidth}
    \centering
    \includegraphics[width=\linewidth]{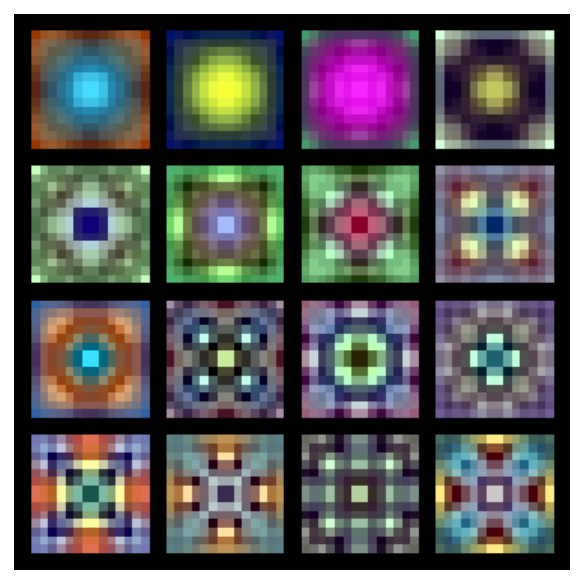}
    \caption{A1}
  \end{subfigure}
  \hspace{0.03\linewidth}
  \begin{subfigure}[b]{0.25\linewidth}
    \centering
    \includegraphics[width=\linewidth]{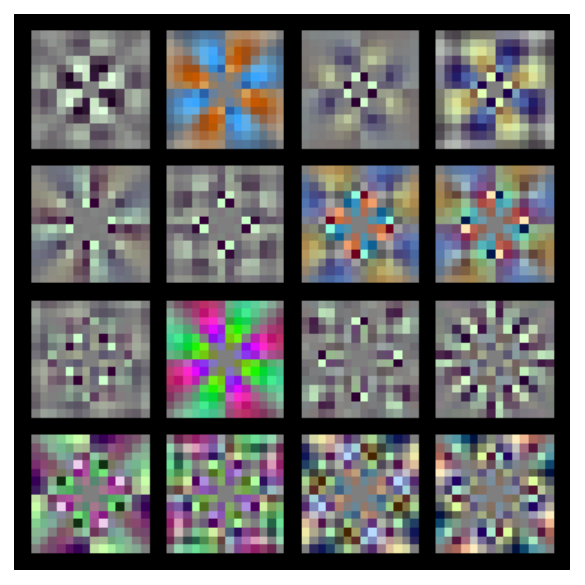}
    \caption{A2}
  \end{subfigure}
  \hspace{0.03\linewidth}
  \begin{subfigure}[b]{0.25\linewidth}
    \centering
    \includegraphics[width=\linewidth]{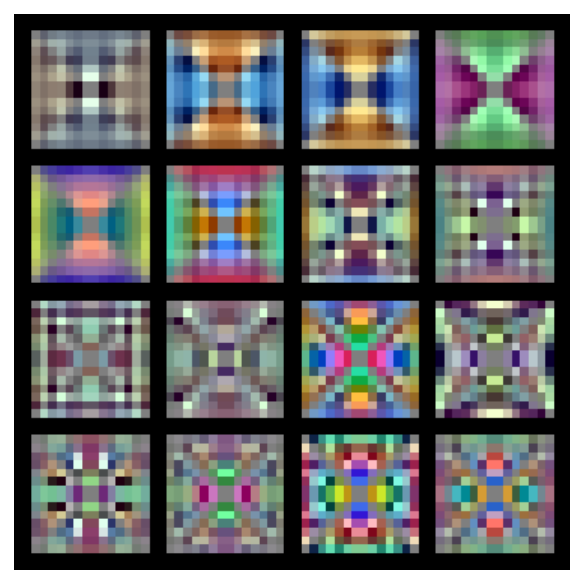}
    \caption{B1}
  \end{subfigure}

  \vspace{0.5em}

  \begin{subfigure}[b]{0.25\linewidth}
    \centering
    \includegraphics[width=\linewidth]{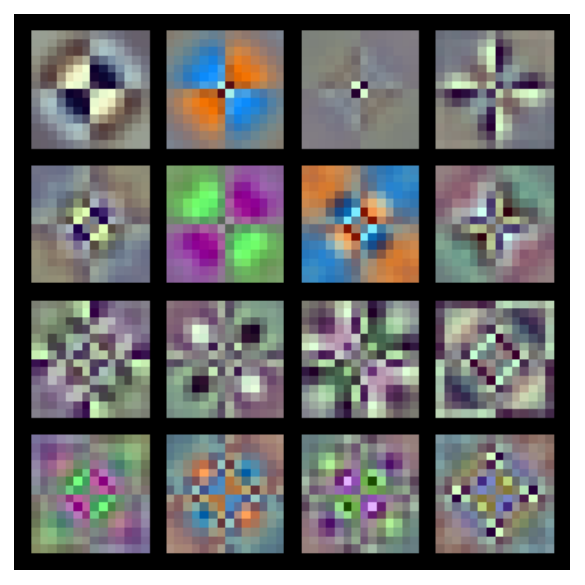}
    \caption{B2}
  \end{subfigure}
  \hspace{0.03\linewidth}
  \begin{subfigure}[b]{0.25\linewidth}
    \centering
    \includegraphics[width=\linewidth]{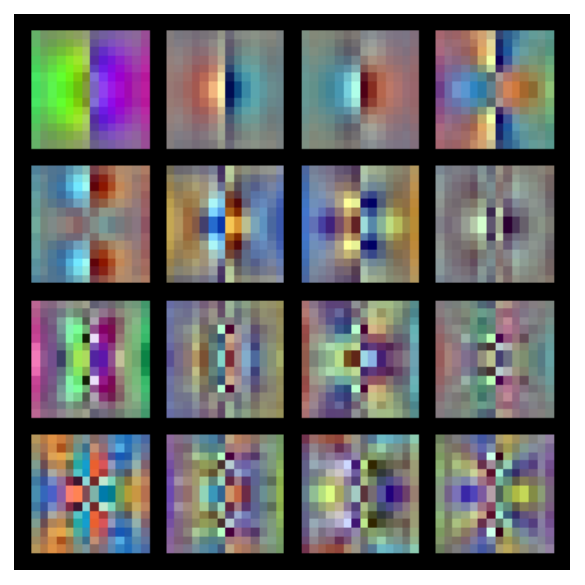}
    \caption{E1}
  \end{subfigure}
  \hspace{0.03\linewidth}
  \begin{subfigure}[b]{0.25\linewidth}
    \centering
    \includegraphics[width=\linewidth]{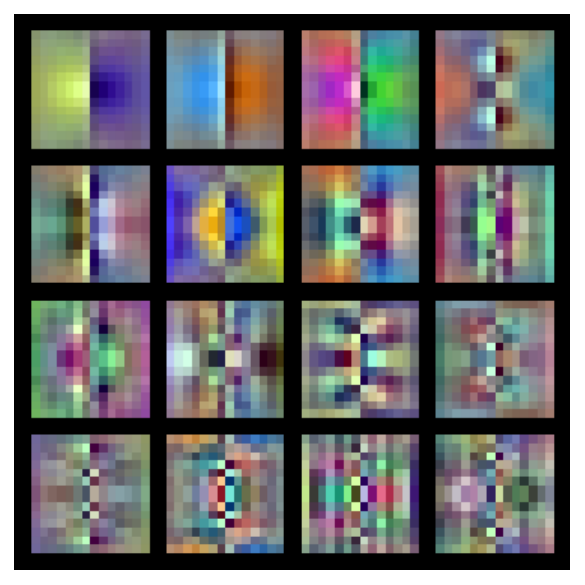}
    \caption{E2}
  \end{subfigure}

  \vspace{1em}

  \caption*{\textbf{(b) Quick ViT DINOv2 Learned Filters}}

  \begin{subfigure}[b]{0.25\linewidth}
    \centering
    \includegraphics[width=\linewidth]{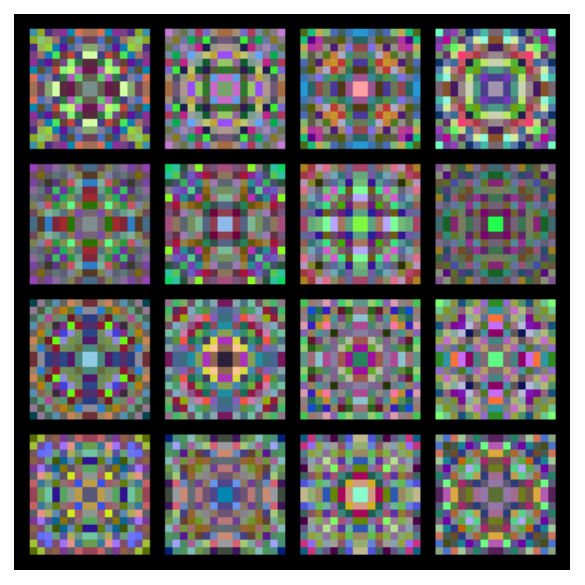}
    \caption{A1}
  \end{subfigure}
  \hspace{0.03\linewidth}
  \begin{subfigure}[b]{0.25\linewidth}
    \centering
    \includegraphics[width=\linewidth]{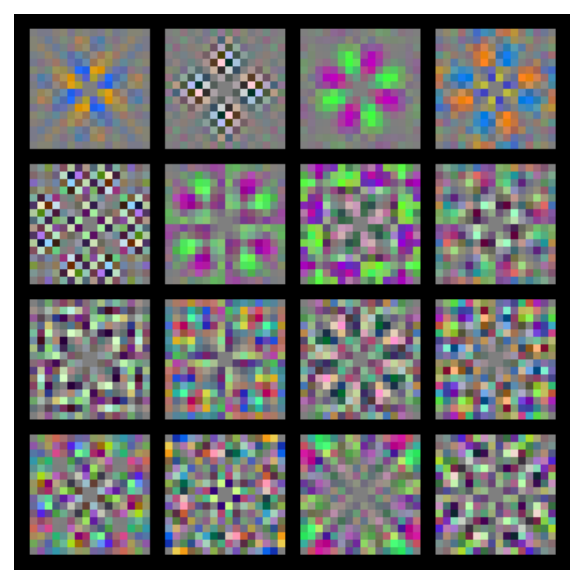}
    \caption{A2}
  \end{subfigure}
  \hspace{0.03\linewidth}
  \begin{subfigure}[b]{0.25\linewidth}
    \centering
    \includegraphics[width=\linewidth]{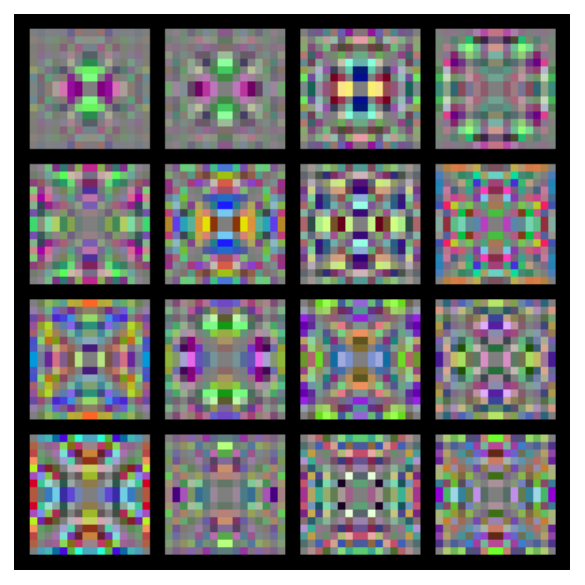}
    \caption{B1}
  \end{subfigure}

  \vspace{0.5em}

  \begin{subfigure}[b]{0.25\linewidth}
    \centering
    \includegraphics[width=\linewidth]{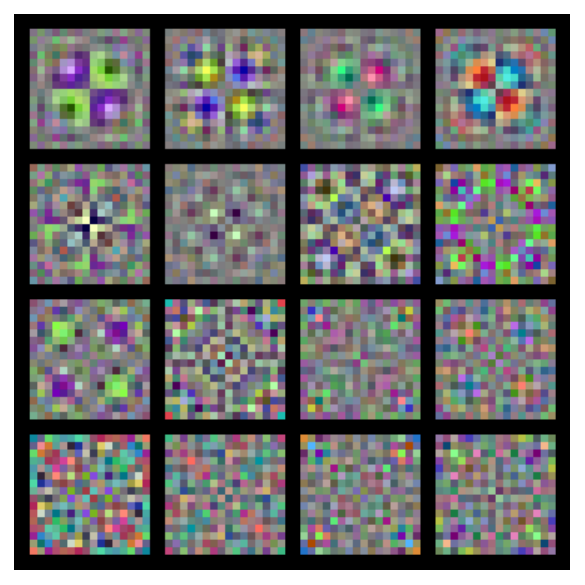}
    \caption{B2}
  \end{subfigure}
  \hspace{0.03\linewidth}
  \begin{subfigure}[b]{0.25\linewidth}
    \centering
    \includegraphics[width=\linewidth]{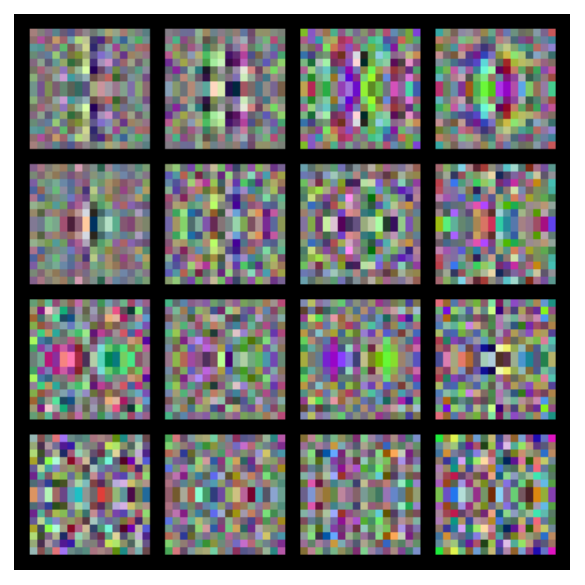}
    \caption{E1}
  \end{subfigure}
  \hspace{0.03\linewidth}
  \begin{subfigure}[b]{0.25\linewidth}
    \centering
    \includegraphics[width=\linewidth]{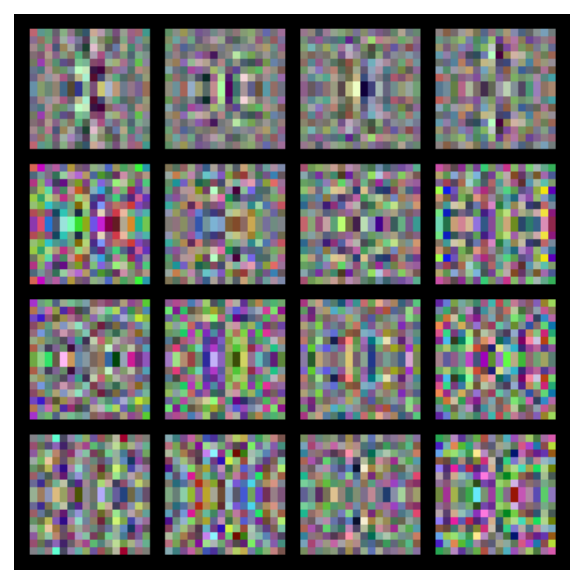}
    \caption{E2}
  \end{subfigure}

  \caption{\textbf{Comparison of learned patch embedding filters.} \textbf{(a)} DeiT III. \textbf{(b)} DINOv2. Each figure shows the top-16 principal components of the Quick Vist PatchEmbed filter for a specific feature type.}
  \label{fig:filters-comparison}
\end{figure}

The PatchEmbed layer contains filters mapping the input from 3 channel dimensions to $C$ embedding dimensions. To illustrate the learned filters, we take inspiration from \cite{dosovitskiy2021an} and visualize the first 16 principal components. In contrast to regular ViTs, we have six different learned filters corresponding to the five irreps. Four for the one-dimensional irrep and two for the E irrep (due to its multiplicity). The results are illustrated in Figure~\ref{fig:filters-comparison}. Interestingly, the learned filters look qualitatively different between the two learning methods. This is similar to how the learned filters of baseline DINOv2 and DeiT III look qualitatively different. DINOv2 training appears to produce more high frequency patterns while DeiT III gives clearer patterns. For the invariant irrep (A1), it appears that the DeiT III training produces a spherical pattern.

\section{Invariantization}\label{appendix:invariantization}

There are multiple options to produce \octicg invariant features, i.e. mapping tokens of type $\rho_\text{chan}=\frac{C}{8}\rho_\text{iso}$ to features of type $C\rho_\text{A1}$ (here denoted in short as \emph{invariantization}).
We let $\psi$ be a function mapping from features of type $\frac{C}{8}\rho_\text{iso}$ to features of type $\frac{KC}{8}\rho_\text{A1}$ for some $K$ which can be larger or smaller than $8$.
These $KC/8$ dimensions are then mapped through a small MLP to $C$ dimensions again.

\subsubsection{Linear Invariant (Linear).} The linear invariant simply extracts the invariant irrep. Here, $K=1$.
\begin{equation}
    \psi(x_\text{A1}, x_\text{A2}, x_\text{B1}, x_\text{B2}, x_\text{E11}, x_\text{E12}, x_\text{E21}, x_\text{E22}) = x_\text{A1}.
\end{equation}

\subsubsection{Triple Correlation (Triple Corr.).} The triple correlation method~\cite{sanborn2023general,kakarala2012bispectrum}
extracts a complete set of third order homogeneous polynomial invariants from a signal over \octicg.
We computed a basis for all third order invariant homogeneous polynomials using Macaulay2~\cite{Macaulay2, ferraro2024invariantring} and use the basis elements as invariant.
Here, $K=15$.
{\tiny
\begin{equation}
    \begin{split}
    \psi(x_\text{A1}&, x_\text{A2}, x_\text{B1}, x_\text{B2}, x_\text{E11}, x_\text{E12}, x_\text{E21}, x_\text{E22})  \\
    = \bigg(&
    x_\text{A1}^3,
    x_\text{A1}(x_\text{E21}^2 + x_\text{E22}^2),
    x_\text{A1}(x_\text{E11}x_\text{E21} + x_\text{E12}x_\text{E22}),
    x_\text{A1}(x_\text{E11}^2 + x_\text{E12}^2),
    x_\text{A1}x_\text{B2}^2,
    x_\text{A1}x_\text{B1}^2,
    x_\text{A1}x_\text{A2}^2,
    \\ &x_\text{B2}x_\text{E21}x_\text{E22},
    x_\text{B2}x_\text{E12}x_\text{E21} + x_\text{B2}x_\text{E11}x_\text{E22},
    x_\text{B2}x_\text{E11}x_\text{E12},
    x_\text{B1}x_\text{E21}^2 - x_\text{B1}x_\text{E22}^2,
    \\ &x_\text{B1}x_\text{E11}x_\text{E21} - x_\text{B1}x_\text{E12}x_\text{E22},
    x_\text{B1}x_\text{E11}^2 - x_\text{B1}x_\text{E12}^2,
    x_\text{A2}x_\text{E12}x_\text{E21} - x_\text{A2}x_\text{E11}x_\text{E22},
    x_\text{A2}x_\text{B1}x_\text{B2} \bigg)
    \end{split}
\end{equation}
}

\subsubsection{Power spectrum.} A common invariant is the power spectrum. We use the following variant, with $K=6$.
\begin{equation}
    \psi(x_\text{A1}, x_\text{A2}, x_\text{B1}, x_\text{B2}, x_\text{E1}, x_\text{E2}) = \left(x_\text{A1}, |x_\text{A2}|, |x_\text{B1}|, |x_\text{B2}|, \|x_\text{E1}\|, \|x_\text{E2}\| \right)
\end{equation}

\subsubsection{Polynomial.} 
Similar to the triple correlation, we can consider a polynomial basis for the full invariant ring. This was computed using Macaulay2, yielding $K=32$.

{\small
\begin{equation}
    \begin{split}
    \psi(x_\text{A1}&, x_\text{A2}, x_\text{B1}, x_\text{B2}, x_\text{E11}, x_\text{E12}, x_\text{E21}, x_\text{E22})  \\
    = \bigg(&
    x_\text{A1},
    x_\text{E21}^2 + x_\text{E22}^2,
    x_\text{E11}x_\text{E21} + x_\text{E12}x_\text{E22},
    x_\text{E11}^2 + x_\text{E12}^2,
    x_\text{B2}^2,
    x_\text{B1}^2,
    x_\text{A2}^2,
    \\ &x_\text{B2}x_\text{E21}x_\text{E22},
    x_\text{B2}x_\text{E12}x_\text{E21} + x_\text{B2}x_\text{E11}x_\text{E22},
    x_\text{B2}x_\text{E11}x_\text{E12},
    x_\text{B1}x_\text{E21}^2 - x_\text{B1}x_\text{E22}^2,
    \\ &x_\text{B1}x_\text{E11}x_\text{E21} - x_\text{B1}x_\text{E12}x_\text{E22},
    x_\text{B1}x_\text{E11}^2 - x_\text{B1}x_\text{E12}^2,
    x_\text{A2}x_\text{E12}x_\text{E21} - x_\text{A2}x_\text{E11}x_\text{E22},
    \\ &x_\text{A2}x_\text{B1}x_\text{B2},
    x_\text{E21}^4 + x_\text{E22}^4,
    x_\text{E11}x_\text{E21}^3 + x_\text{E12}x_\text{E22}^3,
    x_\text{E11}^2x_\text{E21}^2 + x_\text{E12}^2x_\text{E22}^2,
    \\ & x_\text{E11}^3x_\text{E21} + x_\text{E12}^3x_\text{E22},
    x_\text{E11}^4 + x_\text{E12}^4,
    \\ &x_\text{B1}x_\text{B2}x_\text{E12}x_\text{E21} - x_\text{B1}x_\text{B2}x_\text{E11}x_\text{E22},
    x_\text{A2}x_\text{B2}x_\text{E21}^2 - x_\text{A2}x_\text{B2}x_\text{E22}^2,
    \\ & x_\text{A2}x_\text{B2}x_\text{E11}x_\text{E21} - x_\text{A2}x_\text{B2}x_\text{E12}x_\text{E22},
    x_\text{A2}x_\text{B2}x_\text{E11}^2 - x_\text{A2}x_\text{B2}x_\text{E12}^2,
    \\ &x_\text{A2}x_\text{B1}x_\text{E21}x_\text{E22},
    x_\text{A2}x_\text{B1}x_\text{E12}x_\text{E21} + x_\text{A2}x_\text{B1}x_\text{E11}x_\text{E22},
    x_\text{A2}x_\text{B1}x_\text{E11}x_\text{E12},
    \\ & x_\text{A2}x_\text{E21}^3x_\text{E22} - x_\text{A2}x_\text{E21}x_\text{E22}^3,
    x_\text{A2}x_\text{E12}x_\text{E21}^3 - x_\text{A2}x_\text{E11}x_\text{E22}^3,
    \\ &x_\text{A2}x_\text{E11}x_\text{E12}x_\text{E21}^2 - x_\text{A2}x_\text{E11}x_\text{E12}x_\text{E22}^2,
    x_\text{A2}x_\text{E11}^2x_\text{E12}x_\text{E21} - x_\text{A2}x_\text{E11}x_\text{E12}^2x_\text{E22},
    \\ & x_\text{A2}x_\text{E11}^3x_\text{E12} - x_\text{A2}x_\text{E11}x_\text{E12}^3
    \bigg)
    \end{split}
\end{equation}
}

\subsubsection{Max filtering.} We follow~\cite{cahill2024group} and implement a version of their max filtering invariant.
For this, we have a set of $2C$ learnable $C$-dimensional tokens $\mathbf{y} \in \mathbb{R}^{2C \times C}$, and the $2C$ invariants are given by
\begin{equation}
\psi(x) = \oplus_{k=1}^{2C}\max_{g\in \octicg}\langle \mathbf{y}_k, \rho_{\text{chan}}(g)x\rangle.
\end{equation}

\subsubsection{Canonisation.}
Similar to the max filtering approach, we implement a canonisation where we have a single learnable $C$-dimensional reference token $y$ and compute the $C$-dimensional invariant as
\begin{equation}
\psi(x) = \rho_\text{chan}\left(\mathrm{argmax}_{g\in \octicg}\langle y, \rho_{\text{chan}}(g)x\rangle\right)x.
\end{equation}

We conduct a study of the effect of these different invariantization methods.
A priori, max filtering and canonisation should be more expressive than the others as they are the only invariants considered here that are able to preserve the relative phase information coming from different phase in different copies of $\rho_\text{iso}$.
We train $\octicg(\text{ViT-L/16})$ on ImageNet-1K following the DeiT III recipe. The results are presented in Table \ref{tab:ablations}.
The conclusion is that the simple power spectrum invariant works well, and so we select it as our invariantization of choice in the remainder of the experiments.

\begin{table*}[t]
  \centering
    \caption{
    \textbf{Invariantization Ablation.} Comparing classification accuracy using different invariantization methods on ImageNet-1K using the DeiT III training recipe for 400 epochs for \octicg(ViT-L/16). 
  }
  \label{tab:ablations}
  \small{
        \begin{tabular}{lll}
          & Top-1 & Top-5 \\
          \midrule
          \rowcolor{white}
          Linear & 79.2 & 94.7 \\
          Polynomial & 79.4 & 94.8 \\
          Triple Corr. & 76.8 & 93.7 \\
          Canonisation & \bfseries 79.5 & \bfseries 94.9 \\
          Max Filtering &  77.9 & 93.9 \\
          \rowcolor{lightgray}
          Power Spec. & \bfseries 79.5 & \bfseries 94.9 \\
        \end{tabular}
  }
\end{table*}

\section{Experimental Setting}\label{appendix:experiment}

\subsection{DeiT III}

We train for 400 epochs on ImageNet-1K with an effective batch size of 2048 following~\cite{deitiii:2022} and \cite{flopping-for-flops}. We compare to the figures reported in the respective papers and thus only train the \octicg equivariant Quick ViTs. The training recipe includes heavy data augmentation (e.g. cutmix, mixup and color jitter) and uses the deprecated NVIDIA \texttt{Apex} library. Training is done in mixed precision with the \texttt{lamb} optimizer. The most important hyperparameters are summarized in Table~\ref{tab:deit-hyperparams}. For more details, we refer to~\cite{deitiii:2022, flopping-for-flops} and the official repo used for reproduction \url{https://github.com/facebookresearch/deit}.

Out of distribution (OOD) rotation evaluation simply adds a random 90 degree rotation to the validation set of IN1K and computes the classification accuracy on the randomly rotated dataset. Note, as the publicly available weights are trained for 800 epochs (whereas we compare to the figures reported for 400 epochs in the original paper), we compute the OOD $\Delta$ on these weights.

\begin{table*}[t]
  \centering
  \caption{\textbf{Hyperparameters used in experiments.} Collection of the most important hyperparameters used. For the full set we refer to the original implementations.}
  \label{tab:hyperparams-comparison}

  \begin{subtable}[t]{0.48\textwidth}
    \centering
    \caption{DeiT III hyperparameters}
    \label{tab:deit-hyperparams}
      \begin{tabular}{@{\ }l|cc@{\ }}
\toprule
Model & 
ViT-L/16 & 
ViT-H/14 \\
\midrule
Batch size & 
$128\times 16$ & 
$64\times32$ \\
Optimizer &
LAMB &
LAMB  \\
LR      & 
$3\times 10^{-3}$  & 
$3\times 10^{-3}$ \\
LR decay& 
cosine & 
cosine  \\
Weight decay     &
0.02 & 
0.02 \\
Training epochs & 
400 & 
400  \\
Warmup epochs & 
5 & 
5  \\
Stoch. Depth & 
0.4 & 
0.5  \\
Repeated Aug & 
\cmark &
\cmark \\
Gradient Clip. & 
1.0 & 
1.0 \\
Mixup alpha  & 
0.8 &
0.8 \\
Cutmix alpha &
1.0 &
1.0 
 \\
ColorJitter  & 
0.3  &
0.3 
\\
Loss &
BCE & 
BCE 
\\
 \bottomrule
\end{tabular}
  \end{subtable}
  \hfill
  \begin{subtable}[t]{0.48\textwidth}
    \centering
    \caption{DINOv2 hyperparameters}
    \label{tab:dinov2-hyperparams}
      \begin{tabular}{@{\ }l|cc@{\ }}
\toprule
Model & 
ViT-L/16 & 
ViT-H/16 \\
\midrule
Batch size & 
$64\times 16$ & 
$32\times 32$ \\
Optimizer &
AdamW &
AdamW  \\
Base LR      & 
$4\times 10^{-3}$  & 
$4\times 10^{-3}$ \\
Init layer scale& 
$1.10^{-5}$ & 
$1.10^{-5}$  \\
Weight decay     &
0.04 & 
0.04 \\
Training steps & 
125K & 
125K  \\
Warmup steps & 
12.5K & 
12.5K  \\
Drop path & 
0.3 & 
0.3  \\
K & 
65536 &
65536 \\
Gradient Clip. & 
3.0 & 
3.0 \\
Init EMA momentum  & 
0.8 &
0.8 \\
Koleo loss weight &
0.1 &
0.1 
 \\
iBOT sample prob.  & 
0.5  &
0.5 
\\
iBOT mask ratio &
0.1-0.5 & 
0.1-0.5 
\\
 \bottomrule
\end{tabular}
    \label{tab:methodb-hyperparams}
  \end{subtable}
\end{table*}

\subsection{DINOv2} \label{appendix:dinov2}

We closely follow the implementation in the original paper, only modifying to train in BF16 instead of FP16 for greater stability, and follow the same evaluation protocol for classification. Lacking an official reproduction of the segmentation protocol in DINOv2, we opt for the evaluation protocol created by~\cite{CAPI:2025} for semantic segmentation on ADE20K and VOC2012. In contrast to the original DINOv2 paper, we decide to limit our study to ViT-L and ViT-H, the latter of which was not included in the original paper. We opt for the larger patch size of $P=16$ for all our DINOv2 models to save computational resources. Note, this is the reason why we report fewer FLOPs for our DINOv2 ViT-H models than their DeiT III counterpart (which use $P=14$). 

\begin{figure}[htbp]
  \centering
  \begin{subfigure}[b]{0.45\linewidth}
    \centering
    \includegraphics[width=\linewidth]{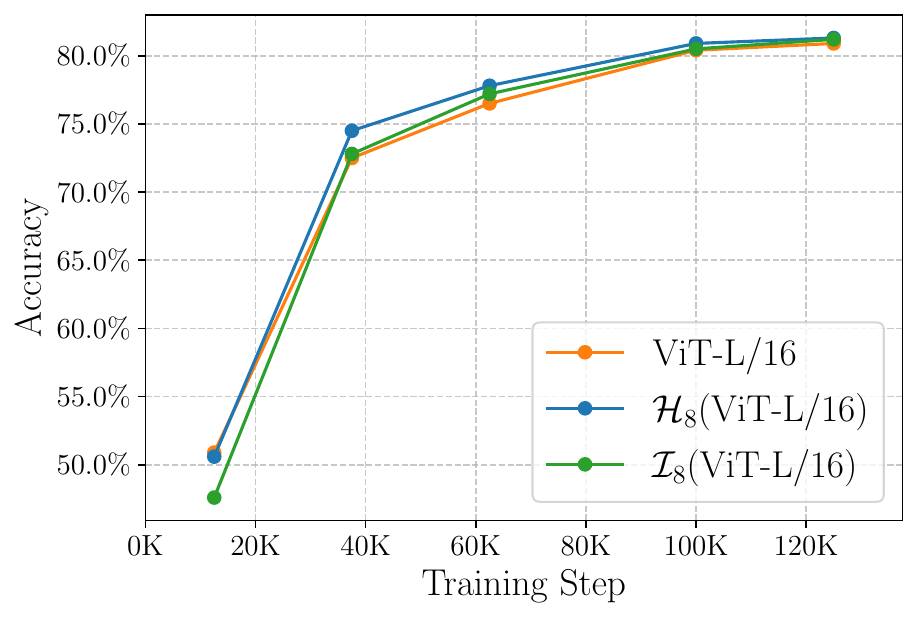}
    \caption{linear}
  \end{subfigure}
  \begin{subfigure}[b]{0.45\linewidth}
    \centering
    \includegraphics[width=\linewidth]{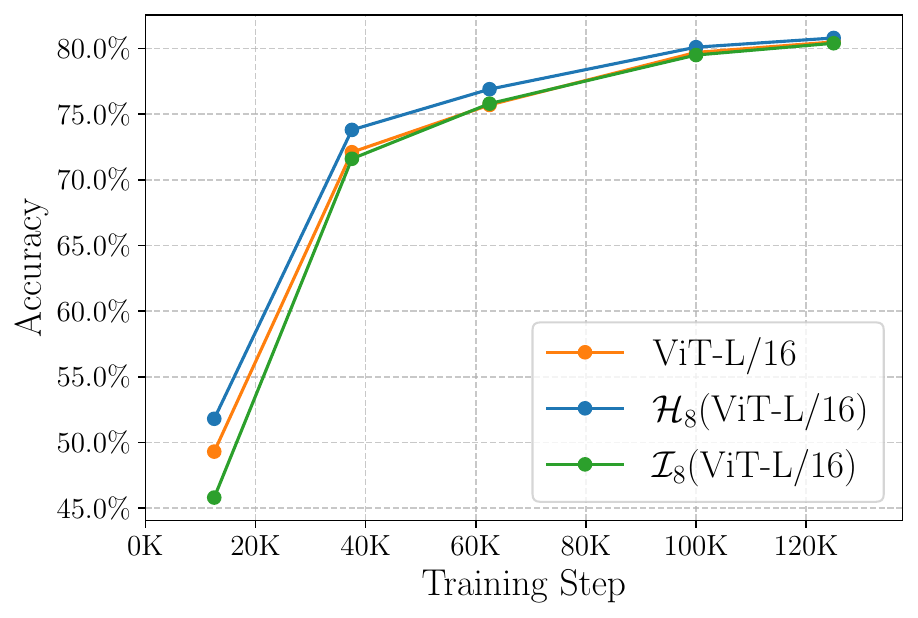}
    \caption{$k$-NN}
  \end{subfigure}
  \caption{
    {\textbf{DINOv2 training progression}.} Classification accuracy development during 125K training steps for linear probe and $k$-NN on frozen features for ViT-L sized models.
  }
  \label{fig:dinov2}
\end{figure}

We train on ImageNet-1K for 125K steps with an effective batch size of 1024 using the \texttt{adamw} optimizer. We train our own baselines for fair comparison (to obtain checkpoints after only training on IN1K). The training progression of the ViT-L/16 family can be visualized in Figure~\ref{fig:dinov2}. The most important hyperparameters are summarized in Table~\ref{tab:dinov2-hyperparams}. For exact details about the configuration of hyper parameters we refer to the base configs in the DINOv2 repo \url{https://github.com/facebookresearch/dinov2}.

We do not implement specific hardware efficient layers for DINOv2 training and instead opt for the standard octic layers that are \texttt{timm} compatible. As such, the octic layers do not leverage \texttt{NestedTensorBlock} and training speedups associated with \texttt{xFormers}. This choice does not impact speed on downstream tasks but slightly decreases pre-training speed. 

We extend the evaluation of DINOv2 (trained on ImageNet-1K) to two popular classification datasets and report the performance in Table~\ref{tab:dinov2-additional-eval}. We find, once again, that our networks achieve similar or better performance than the baseline while using substantially fewer FLOPs.  

\begin{table}[t]
\centering
     \caption{
\textbf{DINOv2 additional evaluation.} We further evaluate the frozen DINOv2 features by classification accuracy on iNaturalist2021~\cite{inat2021} and Places365~\cite{zhou2017places}.
\label{tab:dinov2-additional-eval}}
  \centering
    \begin{tabular}{@{}lcccc@{} c cc c cc @{\ }}
        \toprule 
        Model & FLOPs &&
        \multicolumn{2}{@{}c@{}}{iNaturalist2021 $\uparrow$} && \multicolumn{2}{@{}c@{}}{Places365 $\uparrow$} \\
        \cmidrule{4-5}
        \cmidrule{7-8}
         & ($\times 10^9$) && linear & $k$-NN && linear & $k$-NN \\
        \midrule
        ViT-H/16 & 127.7 && 81.7 & 81.0 && 34.7 & 30.6 \\
        $\mathcal{I}_8(\text{ViT-H/16})$ & 77.7 && 81.9 & 80.9 && 33.9 & 29.2 \\
        $\mathcal{H}_8(\text{ViT-H/16})$ & 77.5 && \bfseries82.2 & \bfseries81.4 && \bfseries35.1 & \bfseries31.1 \\
        \midrule
        ViT-L/16 & 61.9 && 80.9 & 80.5 && 33.2 & 28.4\\
        $\mathcal{I}_8(\text{ViT-L/16})$  & 38.5 && 81.2 & 80.4 && 32.6 & 28.0 \\
        $\mathcal{H}_8(\text{ViT-L/16})$ & 37.7 && \bfseries81.3 & \bfseries80.8 && \bfseries33.6 & \bfseries29.4 \\

        \bottomrule		
      \end{tabular}
\end{table}

\subsection{DinoBloom} \label{appendix:dinobloom}

We investigate the performance of our invariant model on white blood cell classification. We follow the procedure of DinoBloom~\cite{dinobloom:2024} and report our results for ViT-L in Table~\ref{tab:dinobloom}. We find that the invariant model $\mathcal{I}_8(\text{ViT-L/16})$ outperforms the baseline on most evaluated metrics. We tried evaluating on a rotated test set and found negligible change in performance for the baseline (the invariant model inherently has no change in performance, similar to the last column of Table \ref{tab:deit}).

For the details of the experiment, we closely follow the training and evaluation protocol of \cite{dinobloom:2024}. In particular, we finetune our DINOv2 checkpoints for 4K iterations (taking approx. 1 hour on an 8$\times$ A100-40GB node) and evaluate on a hold-out split of the Bone Marrow Cytomorphology (BMC)~\cite{matek2021bone} dataset. However, we limit our finetuning datasets to the datasets presented in Table~\ref{tab:dataset_mix}. Note, we follow the same datasplit of BMC as in DinoBloom and thus also refrain from training on that part.

\begin{table}[htbp] \centering
\small
\setlength{\tabcolsep}{2pt}
\caption{\textbf{Hematology datasets}. Dataset mixture for DinoBloom finetuning.}
\label{tab:dataset_mix}

\begin{tabular}{lcr}
\hline
\specialrule{1.5pt}{0pt}{0pt} 
Datasets                          & Modality            & Images \\ 
\specialrule{1.5pt}{0.5pt}{0.5pt} 
BMC    & Bone Marrow        &  171K \\
AML Hehr    & Blood        &  102K \\
APL    & Blood        &  26K \\
AML Matek    & Blood        &  18K \\
Acevedo    & Blood        &  17K \\
Raabing WBC & Blood        &  10K \\
\specialrule{1.5pt}{0.5pt}{0.5pt} 
\textbf{Total}                        &                     & \textbf{354k} \\
\specialrule{1.5pt}{0pt}{0pt} 
\end{tabular}
\normalsize
\end{table}

\begin{table}[t]
\centering
     \caption{
\textbf{Hematology finetuning.} White blood cell classification performance on BMC dataset with 21 highly imbalanced classes after finetuning on hematology data following DinoBloom.
\label{tab:dinobloom}}
  \centering
  \begin{tabular}{@{}lcccc@{} c ccc@{\ }c ccc@{\ }c ccc@{\ }}
    \toprule 
    Model &&
    \multicolumn{3}{@{}c@{}}{1-$k$-NN} &&
    \multicolumn{3}{@{}c@{}}{20-$k$-NN} &&
    \multicolumn{3}{@{}c@{}}{Lin. probe} \\
    \cmidrule{3-5}
    \cmidrule{7-9}
    \cmidrule{11-13}
     && wF1 & Acc & bAcc && wF1 & Acc & bAcc && wF1 & Acc & bAcc \\
    \midrule
    ViT-L/16  && 78.0 & \bfseries78.0 & 57.6 && 83.1 & 83.6 & 54.9 && 84.6 & 84.7 & \bfseries62.2 \\
    $\mathcal{I}_8(\text{ViT-L/16})$  && \bfseries78.1 & \bfseries78.0 & \bfseries61.3 && \bfseries83.5 & \bfseries83.9 & \bfseries55.7 && \bfseries85.0 & \bfseries85.2 & 61.9 \\
    \bottomrule		
  \end{tabular}
\end{table}

\subsection{Rotation error}
In Table~\ref{tab:rotation-error}, we show the difference in classification performance under 45$^\circ$ rotations. The $\mathcal{I}_8$ models are substantially more robust to rotations.

\begin{table}[t]
\centering
     \caption{\textbf{Classification performance under 45$^\circ$ rotations.} In addition to the 90$^\circ$ rotation error we evaluate in Table~\ref{tab:deit}, we show the difference in classification performance from upright images to 45$^\circ$ rotations.}
  \centering
  \begin{tabular}{@{}lr}
    \toprule 
    Model & $\Delta$ acc. \\
    \midrule	
    ViT-H/16 &  -13.1 \\
    $\mathcal{I}_8(\text{ViT-H/16})$ & -5.4 \\
    $\mathcal{H}_8(\text{ViT-H/16})$ & -13.5  \\
    \midrule
    ViT-L/16 & -13.3 \\
    $\mathcal{I}_8(\text{ViT-L/16})$ & -5.7 \\
    $\mathcal{H}_8(\text{ViT-L/16})$ & -13.6 \\
    \bottomrule		
  \end{tabular}
    \label{tab:rotation-error}
\end{table}

\subsection{General Settings} 

\paragraph{Software versioning.} We utilize the PyTorch \cite{PyTorch:NEURIPS2019_9015} and the timm \cite{rw2019timm} libraries for our experiments. We run the same versioning as our benchmarks. For all other experiments, we use Python 3.11.9 and PyTorch 2.6.0 with CUDA 11.8. 

\paragraph{Model sizes.} The model sizes referred to in the paper adhere to the standard terminology used by~\cite{rw2019timm, dosovitskiy2021an}. If we denote the shape by a tuple of (depth, width, attention heads), ViT-L has shape (24, 1024, 16) and ViT-H has shape (32, 1280, 16). Both use MLP dimension four times the size of the embedding dimension (commonly referred to as MLP ratio).  

\paragraph{Calculating throughput.} Throughput and peak memory are measured on a single A100-80GB GPU with batch size fixed to 64 using \texttt{torch.compile}, FlashAttention~\cite{dao2023flashattention2fasterattentionbetter}, and mixed precision. The throughput only measures forward passes with no gradients. Moreover, we utilize 10 warm-up iterations and then average over 100 runs~\cite{flopping-for-flops}. Peak memory is measured with PyTorch's device memory allocation monitor.

\paragraph{Counting FLOPs.}
We count FLOPs using 
\path{fvcore.nn.FlopCountAnalysis}. FLOPs are normalized
with respect to the batch size (i.e.\ we measure FLOPs/image).
We acknowledge that the term FLOPs often leads to confusion. We adopt the terminology of prior work~\cite{deitiii:2022, flopping-for-flops} and the \texttt{fvcore} library for FLOPs, though, strictly speaking, this refers to MACs (as a factor of two is omitted).

\end{document}